\documentclass[preprint,12pt,authoryear]{elsarticle}

\usepackage{amssymb}

\usepackage{amsmath}
\usepackage{subfig}

\journal{Neural Networks}

\begin{document}

\begin{frontmatter}



\title{Chaotic Phase Synchronization and Desynchronization in an Oscillator Network for Object Selection}


\author{Fabricio~A.~Breve\corref{cor1}\fnref{fn1}}
%
\author{Liang~Zhao\fnref{fn2}}
%
\author{Marcos~G.~Quiles\fnref{fn3}}
%
\author{Elbert~E.~N.~Macau\fnref{fn4}}
%
%
%

\begin{abstract}

Object selection refers to the mechanism of extracting objects of interest while ignoring other objects and background in a given visual scene. It is a fundamental issue for many computer vision and image analysis techniques and it is still a challenging task to artificial visual systems. Chaotic phase synchronization takes place in cases involving almost identical dynamical systems and it means that the phase difference between the systems is kept bounded over the time, while their amplitudes remain chaotic and may be uncorrelated. Instead of complete synchronization, phase synchronization is believed to be a mechanism for neural integration in brain. In this paper, an object selection model is proposed. Oscillators in the network representing the salient object in a given scene are phase synchronized, while no phase synchronization occurs for background objects. In this way, the salient object can be extracted. In this model, a shift mechanism is also introduced to change attention from one object to another. Computer simulations show that the model produces some results similar to those observed in natural vision systems.

\end{abstract}


\end{frontmatter}


\section{Introduction}



Synchronization is characterized by a tendency of two or more dynamical systems to operate in coherent rhythm. It is a common phenomenon observed in science, engineering and social life as diverse as clocks, singing crickets, cardiac pacemakers, firing neurons and applauding audiences \citep{Pikovsky2001}. Since the early 1990s, there has been strongly increasing interest in the synchronization of chaotic systems \citep{PC1990}. In fact, one can distinguish among different types of synchronization for chaotic systems, from complete synchronization, to phase synchronization, lag synchronization, anticipating synchronization, and generalized synchronization. Complete synchronization is defined as the complete coincidence of the trajectories of the coupled individual chaotic systems in the phase space. Mathematically, given state variable vectors $\mathbf{x}$ and $\mathbf{y}$ representing two dynamical systems, they are said to be completely synchronized if $|\mathbf x - \mathbf y| \rightarrow 0$ as $t \rightarrow \infty$. Phase synchronization takes place in cases involving almost identical dynamical systems and it means that the phase difference between the systems is kept bounded over the time, while their amplitudes remain chaotic and may be uncorrelated \citep{Pikovsky2001,Rosenblum1996}. A third type of synchronization is lag synchronization, occurring for a stronger coupling of the oscillating subsystems until they become identical in phases and amplitudes, but shifted in time  \citep{Rosenblum1997}. Anticipating synchronization occurs in unidirectional delayed coupling schemes to produce a collective behavior wherein the output of the response system anticipates the output of the drive one \citep{Voss2000}. Generalized synchronization implies that the trajectory of one subsystem is a function of other ones \citep{Uchida2003}.

Evidence from physiological experiments has been accumulating with strong indication on the existence of synchronous rhythmic activities in different areas of the brain of human beings, cats and monkeys \citep{Eckhorn1988,Engel1991,Fries2001,Gong2003,Grey1989,Kim2007,Murthy1992}. It has been suggested that this neuronal oscillation and synchronization have a role in feature binding and scene segmentation problems. The processing through synchronous oscillations is related to the temporal coding: an object is represented by temporal correlation of firing activities of spatially distributed neurons \citep{Malsburg1981}. In practice, a special form of temporal correlation, called \textit{oscillatory correlation}, has been successfully applied to several computational problems \citep{Wang1995,Malsburg1986,Wang1997,Wang2005}. The oscillatory correlation role can be described as follows: oscillators which represent different features of the same object are synchronized, while oscillators coding different objects are desynchronized \citep{Wang2005}. Complete chaotic synchronization has been implemented in oscillatory correlation models with advantage of unbounded capacity of segmentation \citep{Hansel1992,Zhao2000,Zhao2001}.


The role of synchronization in brain functions has received support from neurobiological findings \citep{Jermakowicz2007}. For example, it has been shown that visual attention is strongly linked with synchronization, in which the coherence among neurons responding to the same stimulus is increased \citep{Fries2001,Kim2007,Buia2006,Niebur1994}. Visual attention is the capacity developed by living systems to select just relevant environmental information. It reduces the combinatorial explosion of information received by sensory system \citep{Shic2007,Tsotsos1992}, identifies
the region of the visual input which reaches certain awareness level (focus of attention), and suppresses irrelevant information \citep{Itti2001,Carota2004}. Thus, a visual selection system should satisfy the following requirements:

\begin{itemize}
    \item Considering a combination of features as input,
    the system must highlight (select) the
    regions of the image (or objects) where the focus of attention is directed;
    \item All other locations (objects) of the visual input must be suppressed
    by the system in order to keep the focus of attention on just
    few of the active objects.
    \item The focus of attention must be shifted to other remaining
    active locations (objects). It means that the system must implement a
    type of habituation system where the presence of a fixed winner
    stimulus must be followed by a progressive diminution of its
    response and allowing other stimuli to become active.
\end{itemize}

There are basically two approaches for computer visual attention models. One is location-based and another is object-based \citep{Wang2005}. The majority of visual attention models are implemented by using WTA (Winner Take All) mechanisms, which are compatible to location-based theory. In this case, a single neuron is activated. As a result, the attention is directed to a point or a small area, but not a whole object or a component of object \citep{Itti1998}. Object-based theories consider objects as the basic units of perception acting as a whole in a competitive process for attention \citep{Desimone1995,Roelfsema1998,OCraven1999,Wang1999NN}. This approach not only receives behavioral and neurophysiological experimental support, which shows that the selection of objects plays a central role in primate vision \citep{Roelfsema1998,Richard2008,Wang2005b}, but also matches well to our everyday experience. The model proposed in this paper is object-based, in which visual attention is delivered to the salient object or component.

Owing to the relation between synchronization and visual attention, some visual attention models have been proposed where the complete synchronization among oscillators is used to represent objects \citep{Wang1999NN,Kazanovich2002,Quiles2007b}. However, the synchronization phenomena observed in real experiments rarely represent a complete synchronization. Thus, other forms of synchronization should be taken into consideration. Particularly, phase synchronization is a model of reciprocal interaction, which is believed to be the key mechanism for neural integration in brain. Direct evidence supporting phase synchrony as a basic mechanism for brain integration has recently been provided by extensive studies of visual binding \citep{Varela2001}.


In this paper we propose a chaotic oscillatory correlation network for object selection \citep{breve2009}. In general, the focus of attention can be several locations or objects, the so-called magic number $4 \pm 1$ of human short-term memory \citep{Cowan2001}. In this paper, we limit our objective to select only one salient object at each time. In contrast to other oscillation-based object selection models \citep{Wang1999NN,Kazanovich2002}, our model is the first to use chaotic phase synchronization. It is based on a network of coupled chaotic R\"ossler oscillators \citep{Rosenblum1996}, which is used to create a selection mechanism where one of several objects is segmented and highlighted (receives the focus of attention). As the system runs, the group of neurons representing the salient object of a visual input is locked in phase, which means each oscillator still produces a unique chaotic trajectory, but with their phases bounded. At the same time, the groups of neurons representing other objects in the scene remain with their phases uncorrelated. After receiving attention, the salient object is inhibited in order to permit other objects to receive the focus of attention. This attention shift process can be seen as a habituation system where the presence of a fixed winner stimulus must be followed by a progressive diminution of its response and allowing other stimulus to become active (focus of attention shifting). In this model, we consider pixel contrast as visual attention clue, i.e., the object with the higher contrast in the scene is considered as the most salient object. This feature is also supported by biological findings which show that the perceptual system might encode contrast of features rather than the absolute level of them \citep{Yantis2005}.


This paper is organized as follows. In section 2, the chaotic phase synchronization  is presented. In section 3 the proposed model is described. In Section 4, the results obtained through simulation of the proposed model applied to synthetic and real images are showed. Finally, in Section 5, conclusions are drawn.

\section{Chaotic Phase Synchronization}\label{theory}

Two oscillators are called phase synchronized if their phase difference $\phi_1 - \phi_2$ is kept bounded while their amplitudes may be completely uncorrelated  \citep{Pikovsky2001}, i.e. $|\phi_1 - \phi_2| < M$, as $t \rightarrow \infty$. Here, the phase $\phi$ of an oscillator is defined as follows,
\begin{equation}
    \phi = \Upsilon(\arctan(y/x))
\end{equation}
where $x$ and $y$ are variables of the oscillator and $\Upsilon$ represents the unwrap operation. Due to the unwrap operation, $\phi$ is always an increasing variable.

\begin{figure}
\centering
\includegraphics[width=8.0cm]{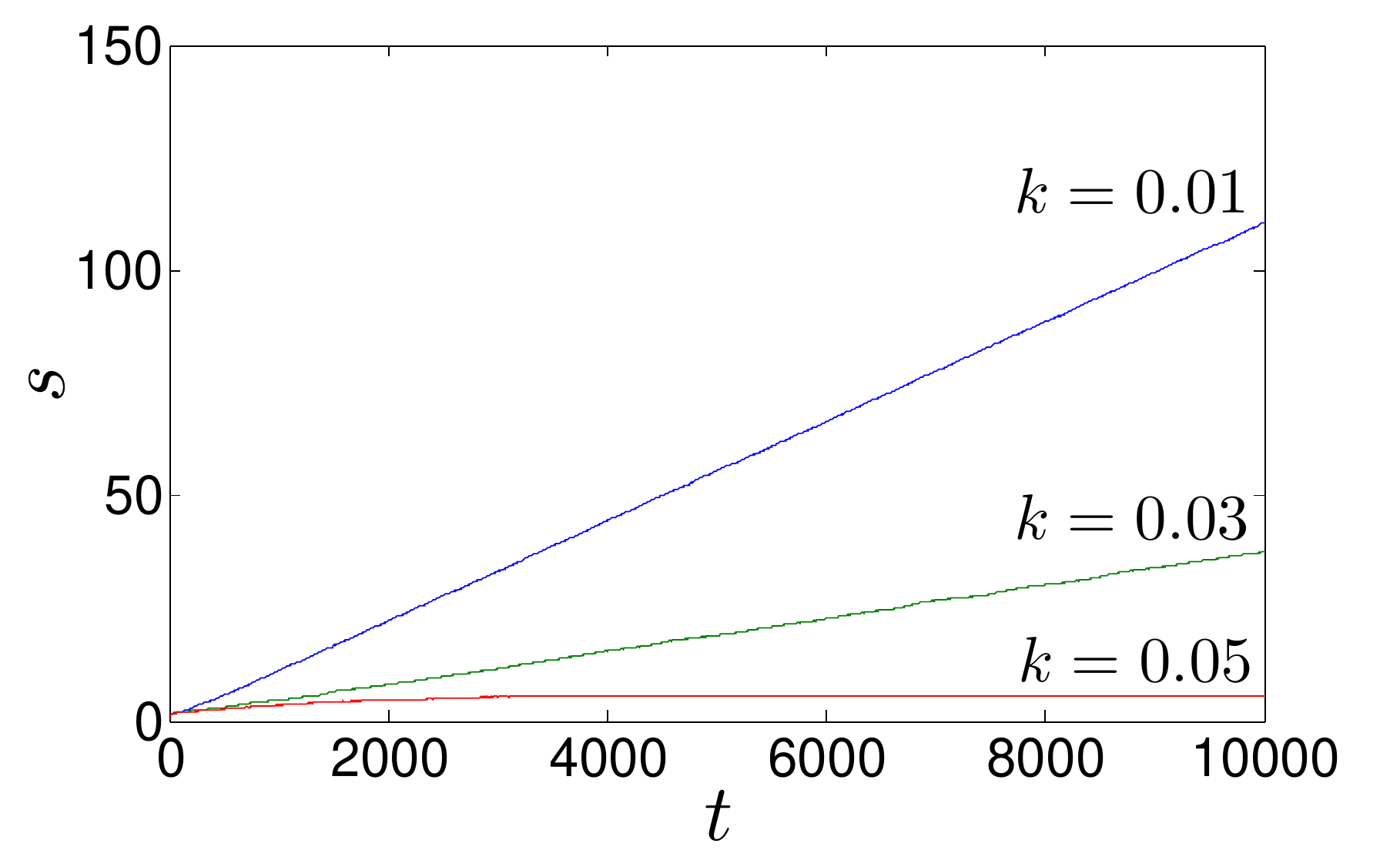}
\caption{Phase standard deviation $s$ of $50$ coupled R\"ossler systems versus time for non-synchronous ($k=0.01$), nearly synchronous ($k=0.03$) and synchronous ($k=0.05$) states. $\omega_{i}$ is randomly chosen in $[0.98, 1.02]$.}
\label{fig:String50}
\end{figure}

In \citep{Rosenblum1996}, it has been shown that two coupled R\"ossler oscillators can be phase synchronized by providing a proper coupling strength. Here we show phase synchronization in an array of $N$ coupled R\"ossler oscillators represented by the following equation:
\begin{align}
     {\dot x}_{i} &= -\omega_{i} y_{i} - z_{i} + k(2x_{i}-x_{i-1}-x_{i+1}), \nonumber \\
     {\dot y}_{i} &= \omega_{i} x_{i} + a y_{i}, \nonumber \\
     {\dot z}_{i} &= b + z_{i} (x_{i} - c),
\end{align}
where parameters $a=0.15$, $b=0.2$ and $c=10$ are held constant with the same values used in \citep{Rosenblum1996}. For each oscillator, $\omega_{i}$ is set randomly in [$0.98 \quad 1.02$] interval. Each oscillator is coupled to its two nearest neighbors, except the two oscillators at each end where each of them is coupled to only one nearest neighbor. Here the amplitude of each R\"ossler oscillator is defined as \citep{Osipov1997}: $A = \sqrt{x^2 + y^2}$. Fig. \ref{fig:String50} shows the transition from unsynchronized state (the standard deviation of phases continuously increases) to synchronized state (the standard deviation of phases does not grow with time) of $50$ locally coupled R\"ossler oscillators as the coupling strength $k$ is increased.

Chaotic phase synchronization has been observed in two-coupled R\"{o}ssler oscillators, a system of $N$ coupled R\"{o}ssler oscillators and other dynamical systems \citep{Pikovsky2001}. Considering a two-coupled R\"{o}ssler oscillators with parameter mismatch, the well-known scenery is: when the coupling strength is zero, no synchronization (including phase synchronization) is observed. In this case, each of the two R\"{o}ssler oscillators has a positive, a negative, and a null (zero) Lyapunov exponent. As the coupling strength is increased, one of the null Lyapunov exponents turns out to be negative. Since the null Lyapunov exponent corresponds to the dynamics of the trajectory, i.e., phase shift of the oscillator, it means that the phases of the two coupled oscillators are correlated. In this case, phase synchronization can be observed, but the amplitudes of the two oscillators remain uncorrelated. As the coupling strength is increased further, complete synchronization (with small difference between trajectories of the two oscillators due to parameter mismatch) occurs. In this case, the system of two coupled R\"{o}ssler oscillators has only one positive, one null and four negative Lyapunov exponents, which means that the two coupled oscillators turn out to be only one chaotic oscillator, i.e., they are synchronized. This scenery also applies to a system of $N$ coupled R\"{o}ssler oscillators. Another factor may change synchronization is the dispersion of parameters. If there isn't parameter mismatch or the parameter mismatch is very small, phase synchronization appears already for vanishing coupling. If the parameter mismatch is small, the coupling strength required for phase synchronization is small, while if the parameter mismatch is large, phase synchronization requires a larger coupling strength. In the last case, it can also occur that the $N$ oscillators are spitted into some clusters.


\section{Model Description}
\label{ModelDescription}

The proposed model is a two dimensional network of R\"ossler Oscillators and it is governed by the following equations:
\begin{align}\label{eq:RosslerGrid}
     {\dot x}_{i,j} &= -\omega_{i,j} y_{i,j} - z_{i,j} + k_{i,j}^+ \Delta^{+} x_{i,j} + k_{i,j}^- \Delta^{-} x_{i,j}, \nonumber \\
     {\dot y}_{i,j} &= \omega_{i,j} x_{i,j} + a y_{i,j}, \nonumber \\
     {\dot z}_{i,j} &= b + z_{i,j} (x_{i,j} - c).
\end{align}
where $(i,j)$ is a lattice point with $1\leq \,i\,\leq N$, $1\,\leq j\,\leq M$. $k_{i,j}^+ \in [0, k_{\max}^+]$ and $k_{i,j}^- \in [0, k_{\max}^-]$ are the positive and negative coupling strength, respectively. They are set according to the pixel contrast, as described below. $\omega_{i,j}$ is also used to code pixel $(i,j)$ contrast, as it will be explained later. $k_{\max}^+$ and $k_{\max}^-$ are set according to the scene. $\Delta^{+} x_{i,j}$ and $\Delta^{-} x_{i,j}$ are positive and negative coupling terms respectively. They are defined by:
\begin{eqnarray}
\Delta^{\pm} x_{i,j}    & = & {\gamma}_{i-1, j-1; i, j}(x_{i-1, j-1} - x_{i, j}) + \nonumber\\
                    &   & {\gamma}_{i-1, j; i, j}(x_{i-1, j} - x_{i, j}) + \nonumber\\
                    &   & {\gamma}_{i-1, j+1; i, j}(x_{i-1, j+1} - x_{i, j}) + \nonumber\\
                    &   & {\gamma}_{i, j-1; i, j}(x_{i, j-1} - x_{i, j}) + \nonumber\\
                    &   & {\gamma}_{i, j+1; i, j}(x_{i, j+1} - x_{i, j}) + \nonumber \\
                    &   & {\gamma}_{i+1, j-1; i, j}(x_{i+1, j-1} - x_{i, j}) + \nonumber\\
                    &   & {\gamma}_{i+1, j; i, j}(x_{i+1, j} - x_{i, j}) + \nonumber\\
                    &   & {\gamma}_{i+1,j+1; i, j}(x_{i+1, j+1} - x_{i, j})
\end{eqnarray}
where
\begin{equation}
{\gamma}_{i, j; p, q} = \left\{
\begin{array}{rcl}
     1, & & \mbox{if oscillator $(i, j)$ is coupled to $(p, q)$}, \\
     0, & & \mbox{otherwise}.
\end{array}\right.
\end{equation}

The positive connections in $\Delta^{+}$ between pairs of neighboring oscillators with similar color will be maintained, while those connections between oscillators of very different color values will be cut. The negative connections in $\Delta^{-}$ are always on, which means that each oscillator is always connected to their $8$-nearest-neighbors, except, of course, for the border oscillators which have fewer neighbors.

The segmentation and object selection scheme is described below. Given an input image, the network is organized so that each oscillator represents a pixel of the image, which means that each oscillator receives a stimulation from its corresponding pixel in the image. In this model, the stimulation is represented by relative pixel contrast $R_{i,j}$. In order to determine relative contrast, we first calculate the absolute contrast $C_{i,j}$ for each pixel by the following equation:
\begin{align}
    C_{i,j} = \frac{\sum_{d}{w^{d} |F^{d}_{i,j}} - F^{d}_{\textrm{avg}}|}{\sum_{d}{w^{d}}} ,
\end{align}
where $(i,j)$ is the pixel index, $F^{d}_{i,j}$ is the feature $d$ value of pixel $(i,j)$ normalized in the interval $[0, 1]$, $w^{d}$ is a weight of feature $d$, and $F^{d}_{\textrm{avg}}$ is the mean value for feature $d$, which is given by:
\begin{align}
    F^{d}_{\textrm{avg}} = \frac{1}{NM} \sum_{i=1}^{i=N} \sum_{j=1}^{j=M} F^{d}_{i,j}.
\end{align}
In this work $4$ features are used, $F^{I}$, $F^{R}$, $F^{G}$ and $F^{B}$, which correspond to the values of intensity (I), red (R), green (G) and blue (B) components from each pixel respectively. The weights are set to $w^I=3$, $w^R=1$, $w^G=1$, and $w^B=1$.

The relative contrast is defined as follows
\begin{align} \label{fixed_contrast}
    R_{i,j} = \exp \Biggl ( -\frac{(1 - C_{i,j})^2}{2\sigma^2} \Biggr).
\end{align}

Once the relative contrast is obtained, it is used to model the oscillator's parameters, so that the oscillators whose correspond to the most salient object (with highest relative contrast) will be synchronized by stronger positive coupling $k_{i,j}^+$ and weaker negative coupling $k_{i,j}^-$. At the same time, those oscillators which correspond to non-salient objects will receive weaker positive coupling and strong negative coupling, thus, they will not be synchronized. Specifically, the parameters are modified by the following rules:

\begin{align}\label{Eq:positive_coupling}
    k_{i,j}^+ = k_{\max}^+ R_{i,j},
\end{align}

\begin{align}\label{Eq:negative_coupling}
    k_{i,j}^- = k_{\max}^- (1 - R_{i,j}),
\end{align}

\begin{align}\label{Eq:omega}
    \omega_{i,j} = 1 - \frac{\Delta_\omega}{2} +  \Delta_\omega C_{i,j},
\end{align}
where $\omega_{i,j}$ varies in $[1 - \frac{\Delta\omega}{2}, 1 + \frac{\Delta\omega}{2}]$ interval and $\Delta\omega$ is set according to the scene.

A new finding presented in this paper is that negative coupling strength does not help but destroys phase synchronization. At the oscillators corresponding to pixels which have the highest contrast, the negative coupling strength tends to zero ($k_{i,j}^- \rightarrow 0$), thus the objects formed by high contrast pixels are nearly unaffected by negative strengths, at the same time that the positive strengths keep their oscillators synchronized in phase. Meanwhile, at the oscillators corresponding to pixels which have less contrast, the negative coupling strength is higher ($k_{i,j}^- \rightarrow k_{\max}^-$), thus these oscillators will repel each other. Finally, only the oscillators corresponding to the salient object will remain with their trajectories synchronized in phase while the other objects will have trajectories with different phases. These features satisfy the essential requirements of a visual selection system described above.

In this paper, the theoretical results presented in Sec. \ref{theory} has been applied in our model. Specifically, the parameters of each R\"{o}ssler oscillator are set as the same values used in \citep{Rosenblum1996}. Thus, each R\"{o}ssler oscillator presents a phase-coherent chaotic attractor \citep{Zhao2004,Zhao2005}. The coupling structure is similar to the model presented in \citep{Osipov1997}, i.e., each oscillator in the network is coupled to some of its eight nearest neighbors with similar color values. We also assume free-end boundary condition \citep{Osipov1997}. The distribution of parameter $\omega_i$ is controlled to be small. Besides of these configurations, each oscillator representing a high contrast pixel (part of the salient object in the input image) receives a strong positive coupling strength by Eq. \ref{Eq:positive_coupling}, a weak negative coupling strength by Eq. \ref{Eq:negative_coupling}, and a relative high oscillating frequency by Eq. \ref{Eq:omega}. On the contrary, each oscillator representing a low contrast pixel (part of the background objects), receives a weak positive coupling strength by Eq. \ref{Eq:positive_coupling}, a strong negative coupling strength by Eq. \ref{Eq:negative_coupling} and a relative low oscillating frequency by Eq. \ref{Eq:omega}. With these configurations at hand and based on the theoretical results presented in Sec. \ref{theory}, it is expected that chaotic phase synchronization occurs among the group of oscillators representing the salient object due to the relative strong positive and weak negative coupling, while no phase synchronization is observed among oscillators representing other objects because of the relative strong negative and weak positive coupling.

From our everyday experience, we know that we cannot hold attention to an object for long time, i.e., the focus of attention must be shifted to other remaining
active objects. This shift mechanism can be implemented in our model by using Eq. \ref{moving_contrast} in place of Eq. \ref{fixed_contrast}. In this case, all other equations are maintained.
\begin{align}\label{moving_contrast}
    R_{i,j} = \exp \Biggl ( -\frac{(t/t_{end} - C_{i,j})^2}{2\sigma^2} \Biggr),
\end{align}
where $t_{end}$ is the whole time of simulation. Eq. \ref{fixed_contrast} defines fixed contrast, i.e., once an object with the highest contrast wins, it stays in the focus of attention forever; while Eq. \ref{moving_contrast} defines moving contrast, i.e., the system will highlight various objects with different levels of contrast. Eq. \ref{moving_contrast} is really a convolution between absolute contrast $C$ and a time shifting Gaussian function.

\section{Computer Simulations}

In this section, we present the simulation results of visual selection tasks by using the proposed model on synthetic and real images. We consider the salient object to be the one which has the largest intensity and color contrast to the other parts of the image. This assumption receives direct support from biological experiments, which show that feature contrast is more important than absolute value of features in visual searching tasks performed by biological visual systems \citep{Wolfe2004,Yantis2005}. The following parameters were held constant for all the experiments: $k^{+}_{max}=0.05$ and $k^{-}_{max}=0.02$.

\begin{figure}
\centering
\subfloat[]
{\includegraphics[width = 8.0cm]{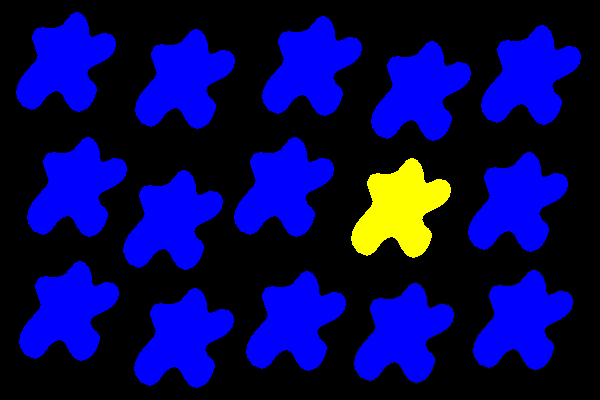}\label{fig:Artif01src}} \\
\subfloat[]
{\includegraphics[width = 8.0cm]{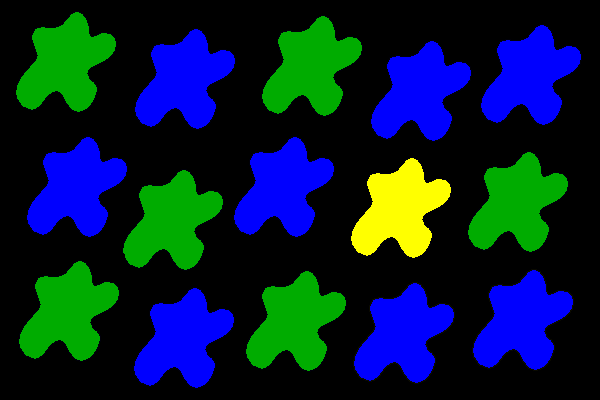}\label{fig:Artif02src}} \\
\subfloat[]
{\includegraphics[width = 8.0cm]{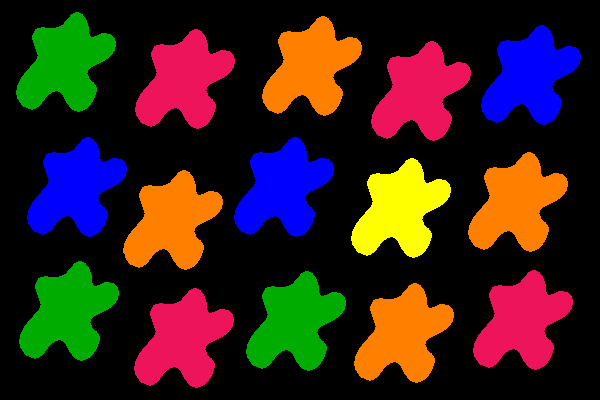}\label{fig:Artif03src}}
\caption{Artificial images with $15$ objects each and different contrast levels: (a) high contrast (b) medium contrast (c) low contrast}
\label{fig:Artifsrc}
\end{figure}

\begin{figure}
\centering
\subfloat[]
{\includegraphics[width = 8.0cm]{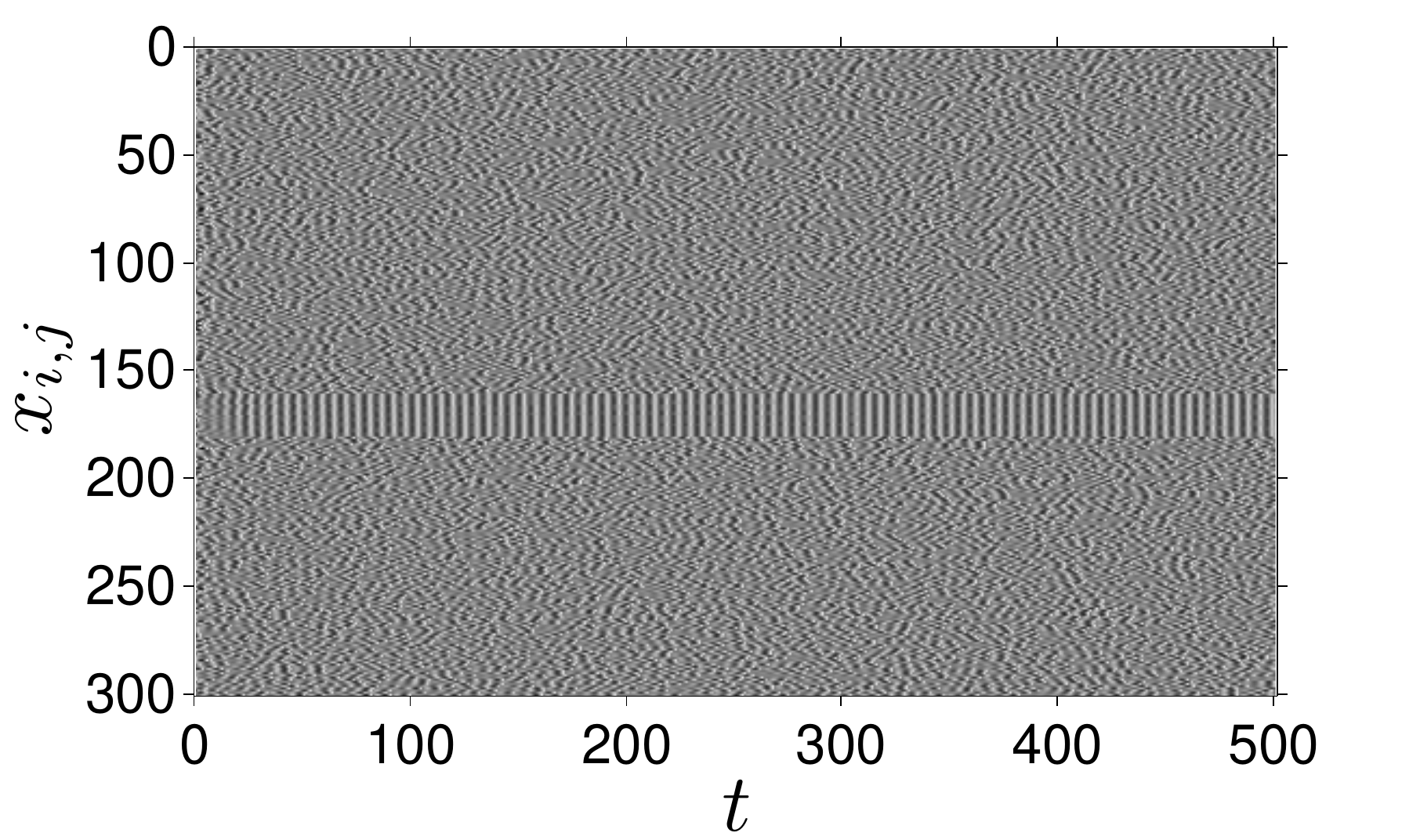}\label{fig:Artif01x}} \\
\subfloat[]
{\includegraphics[width = 8.0cm]{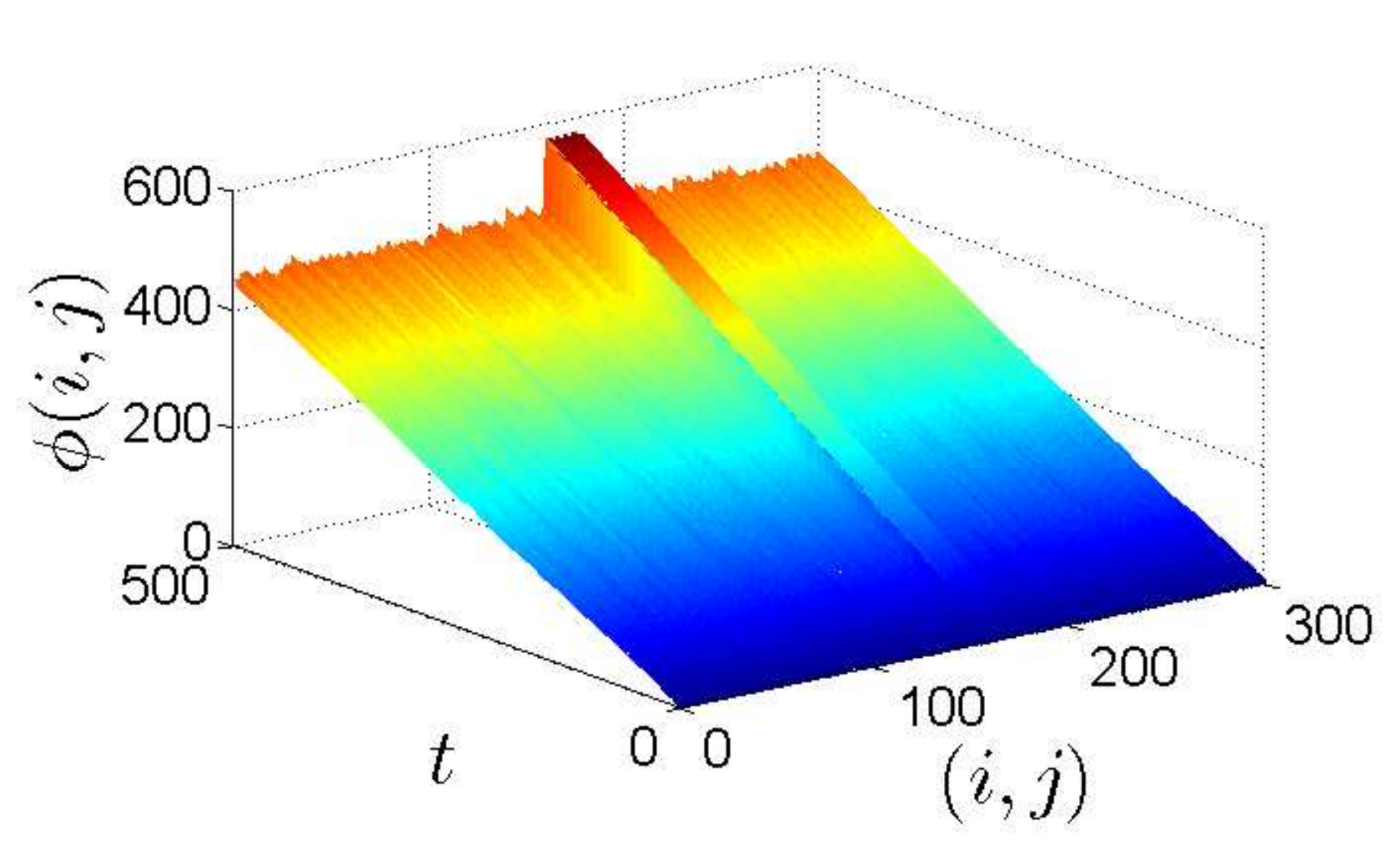}\label{fig:Artif01phase}}
\subfloat[]
{\includegraphics[width = 8.0cm]{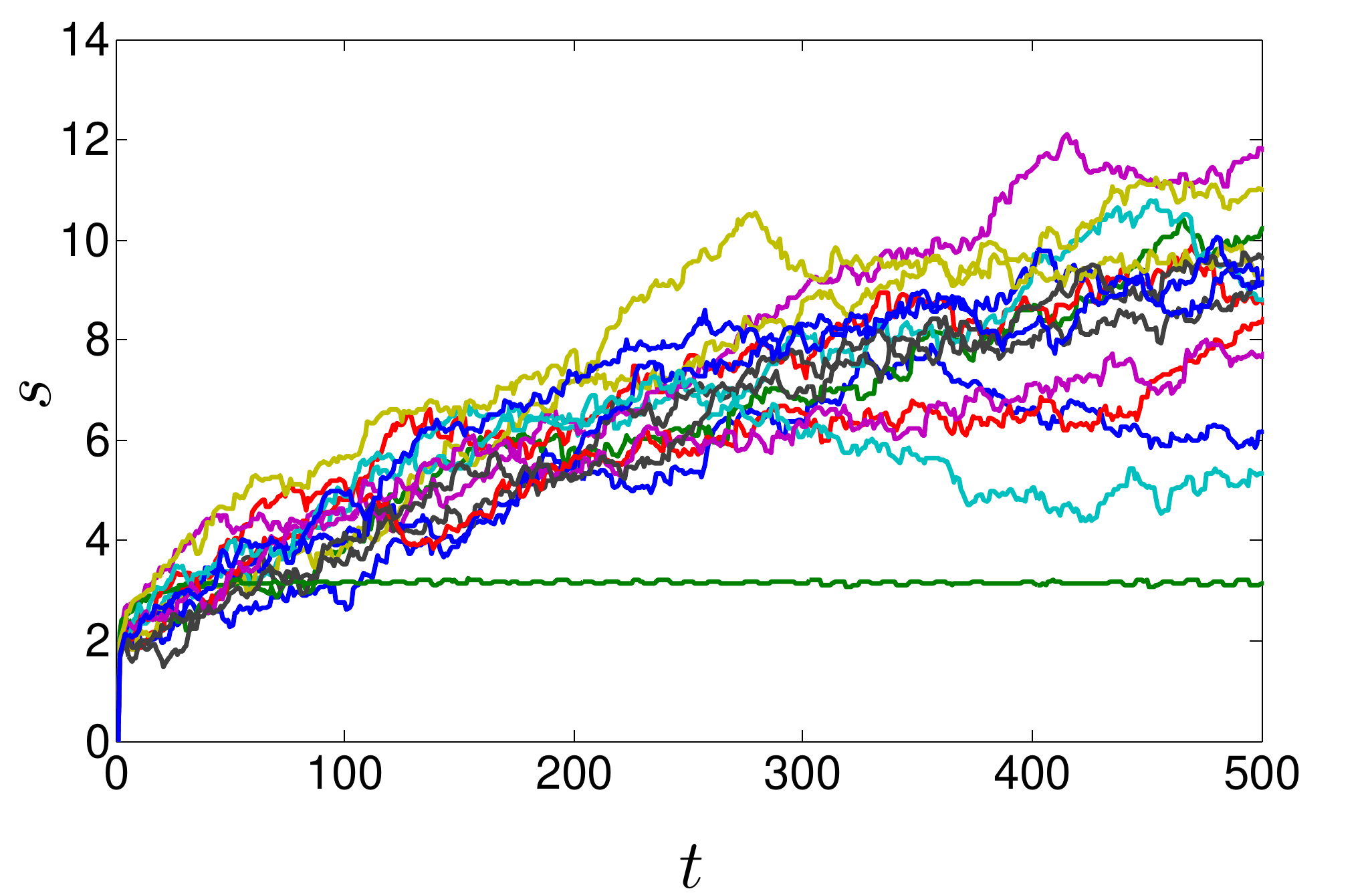}\label{fig:Artif01stdphase}} \\
\subfloat[]
{\includegraphics[width = 8.0cm]{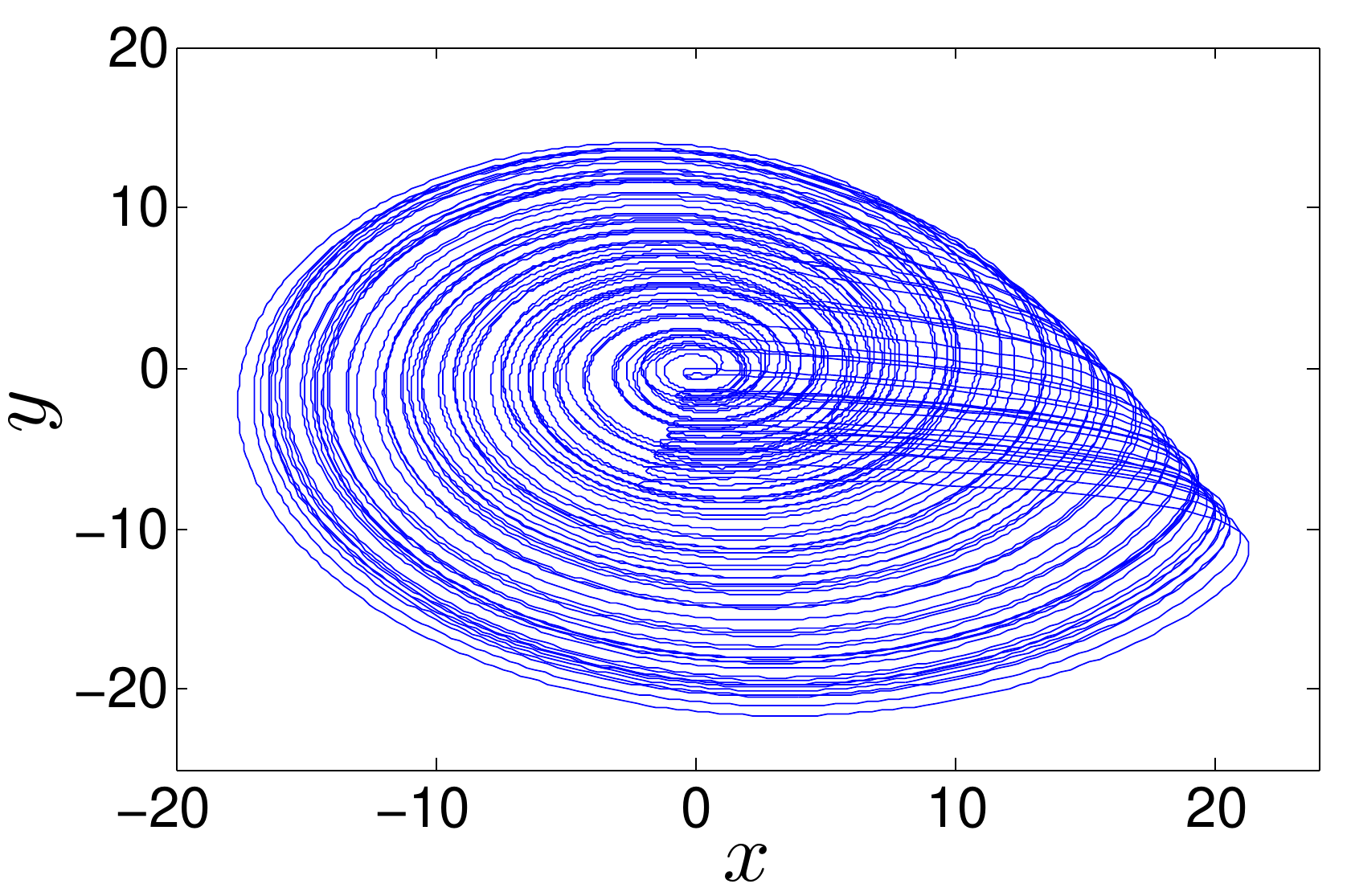}\label{fig:Artif01xynonsync}}
\subfloat[]
{\includegraphics[width = 8.0cm]{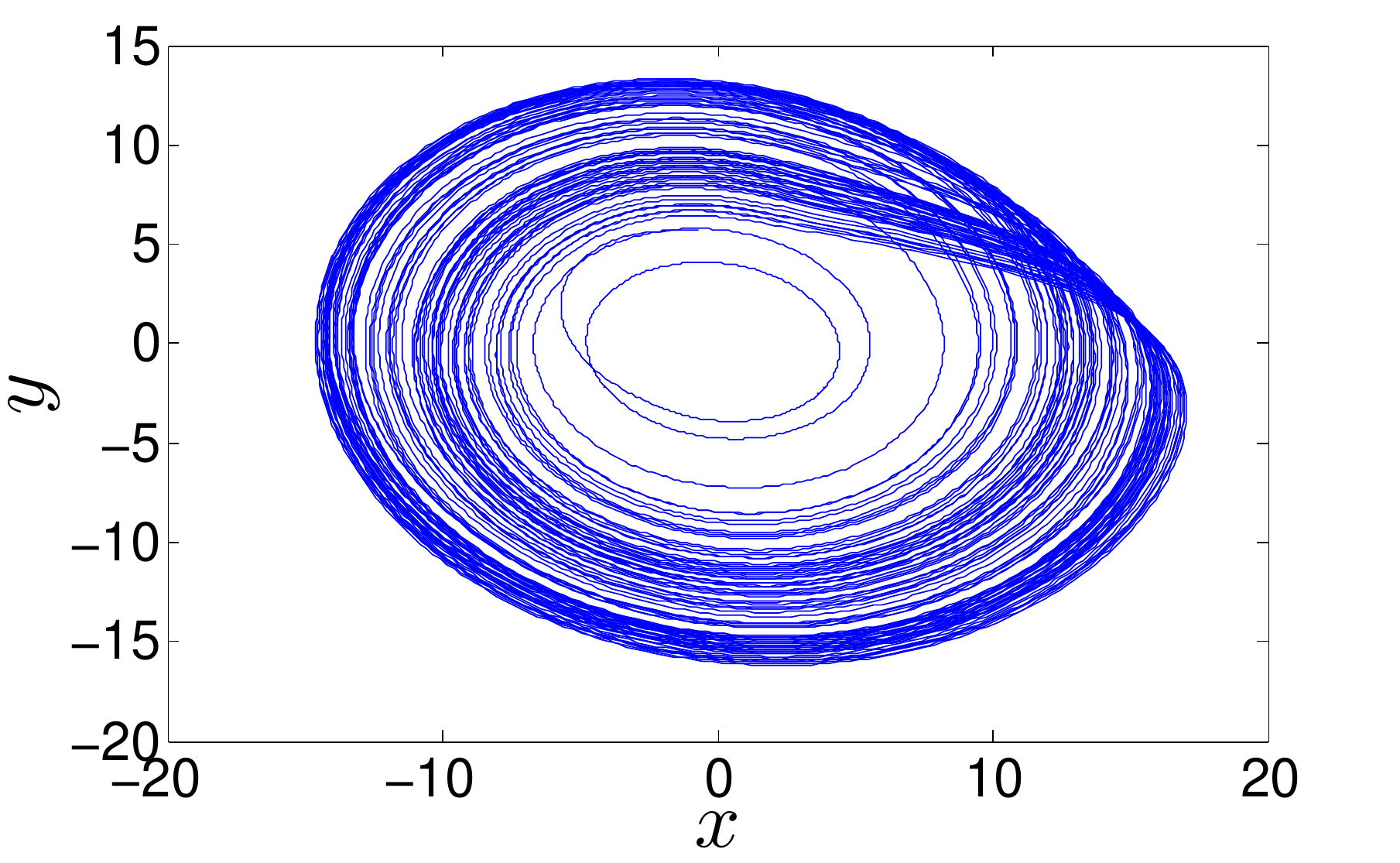}\label{fig:Artif01xysync}}
\caption{Results of artificial image with low contrast. (a) Oscillators behavior. (b) Phase growth. (c) Time series of phase standard deviation of each object.
(d) Phase portrait of a randomly selected oscillator in background objects. (e) Phase portrait of a randomly selected oscillator in salient object.}
\end{figure}

We first perform simulations on the three artificial images shown by Fig. \ref{fig:Artifsrc} with increasing visual complexity. The first experiment is carried out on the image shown by Fig. \ref{fig:Artif01src}, which has $15$ objects, $14$ of them are \textit{blue} and only one is \textit{yellow} (located in second line, forth column), thus becoming the salient object. The free parameters were set as follows: $\sigma = 0.4$ and $\Delta\omega = 0.2$. Figure \ref{fig:Artif01x} shows the behavior of $20$ randomly chosen oscillators (pixels) from each object, where each line corresponds to an oscillator. Dark colors represent the lowest values for the component $x$ of the corresponding oscillator, while bright colors represent the higher values. From lines $161$ to $180$ we can see that the oscillators corresponding to the salient object are phase synchronized (see the formed pattern), while the rest of the oscillators have their phases uncorrelated. Figure \ref{fig:Artif01phase} shows the oscillators phase growth along time, and it can be seen that the oscillators corresponding to the salient object have faster phase growth and also form a plane platform, while the other oscillators have slower phase growth and form an irregular pattern. Figure \ref{fig:Artif01stdphase} shows that the phase standard deviation of the salient object does not increase, indicating that phase synchronization occurs among oscillators representing the salient object, while the phase standard deviations of other objects increase continuously, indicating that no phase synchronization happens. Figure \ref{fig:Artif01xynonsync} and \ref{fig:Artif01xysync} show phase portrait of a randomly selected oscillator in background objects and in salient object, respectively. From these figures, we see that the oscillators are still chaotic after the synchronization process settles down.

\begin{figure}
\centering
\subfloat[]
{\includegraphics[width = 8.0cm]{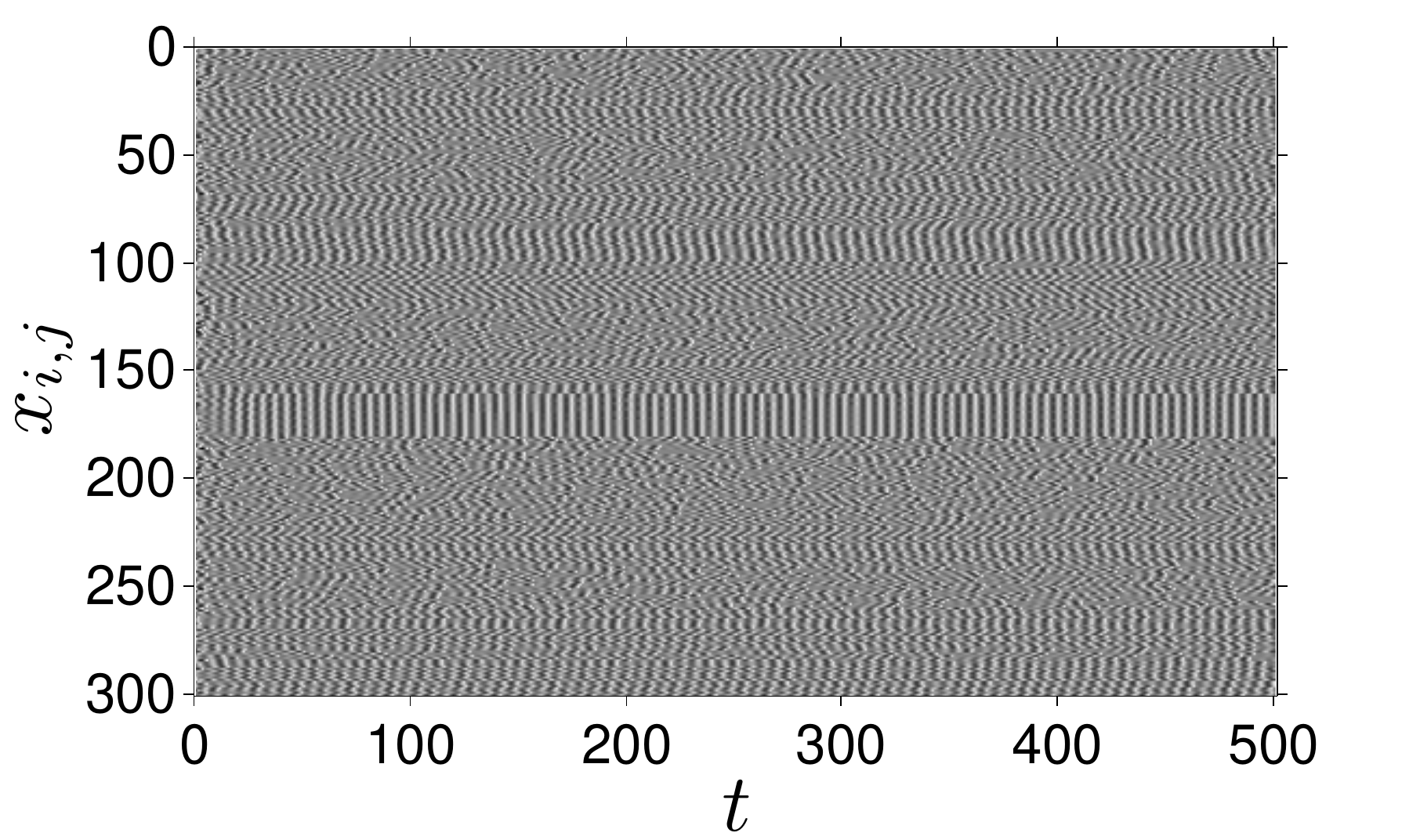}\label{fig:Artif02x}}
\subfloat[]
{\includegraphics[width = 8.0cm]{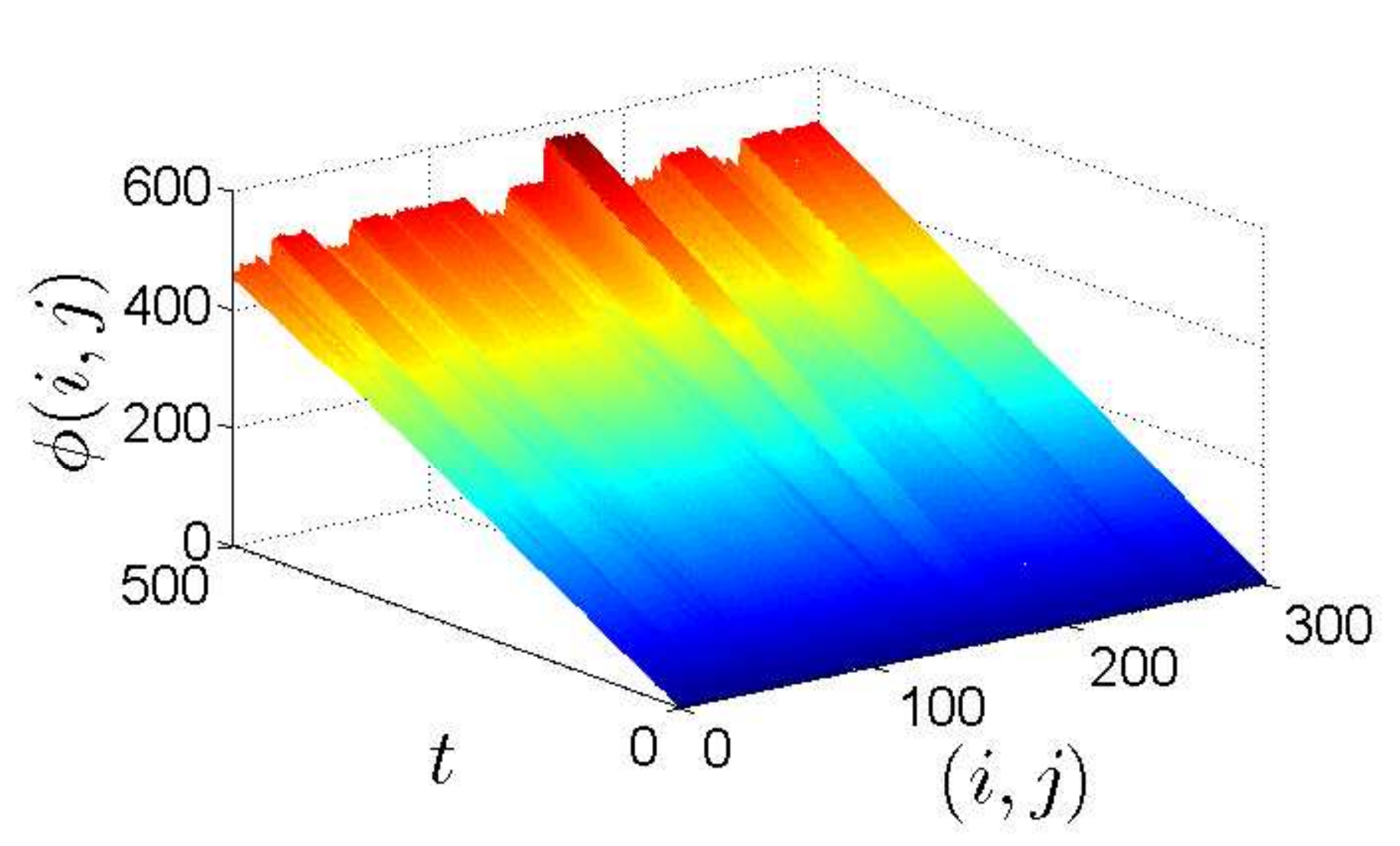}\label{fig:Artif02phase}} \\
\subfloat[]
{\includegraphics[width = 8.0cm]{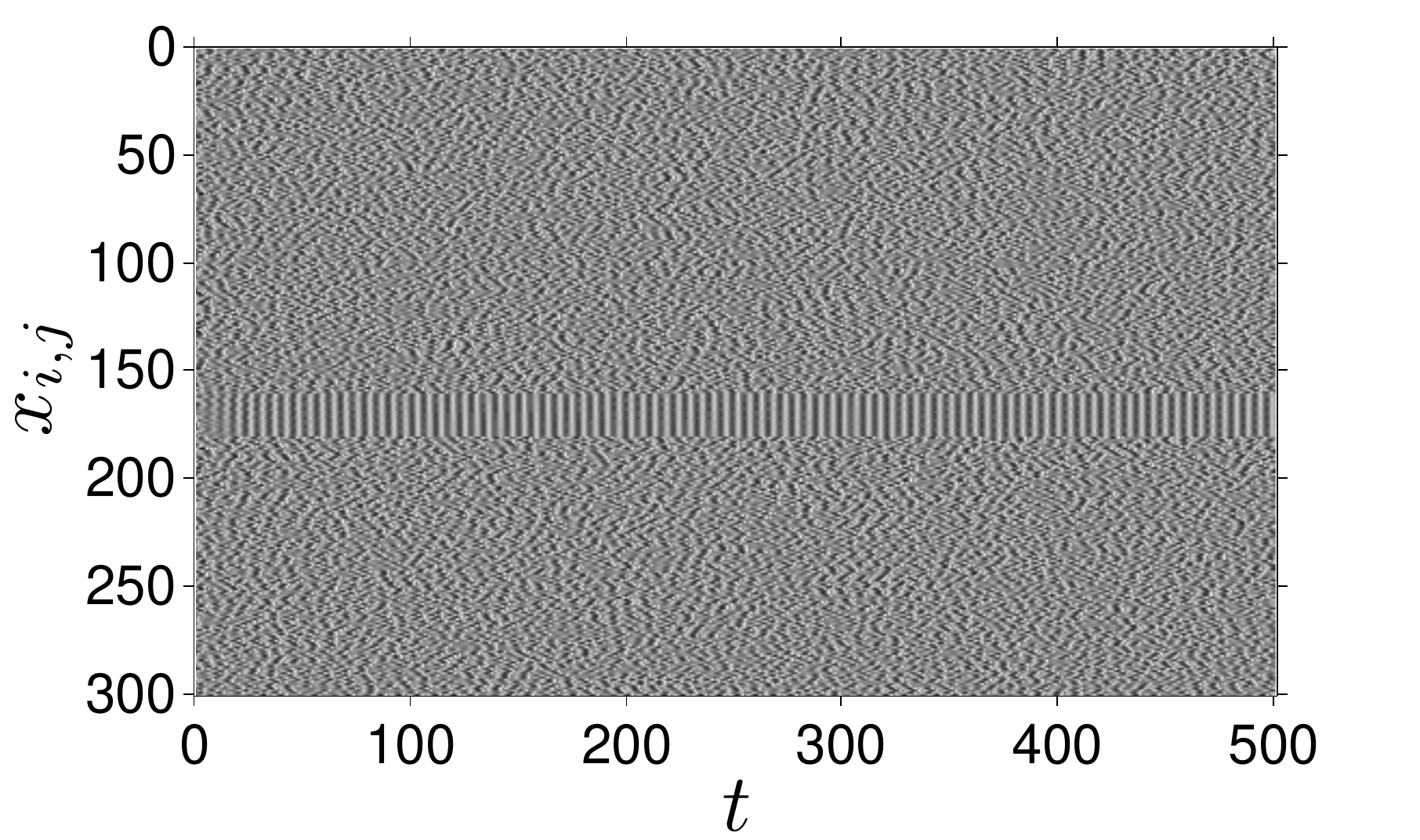}\label{fig:Artif02sx}}
\subfloat[]
{\includegraphics[width = 8.0cm]{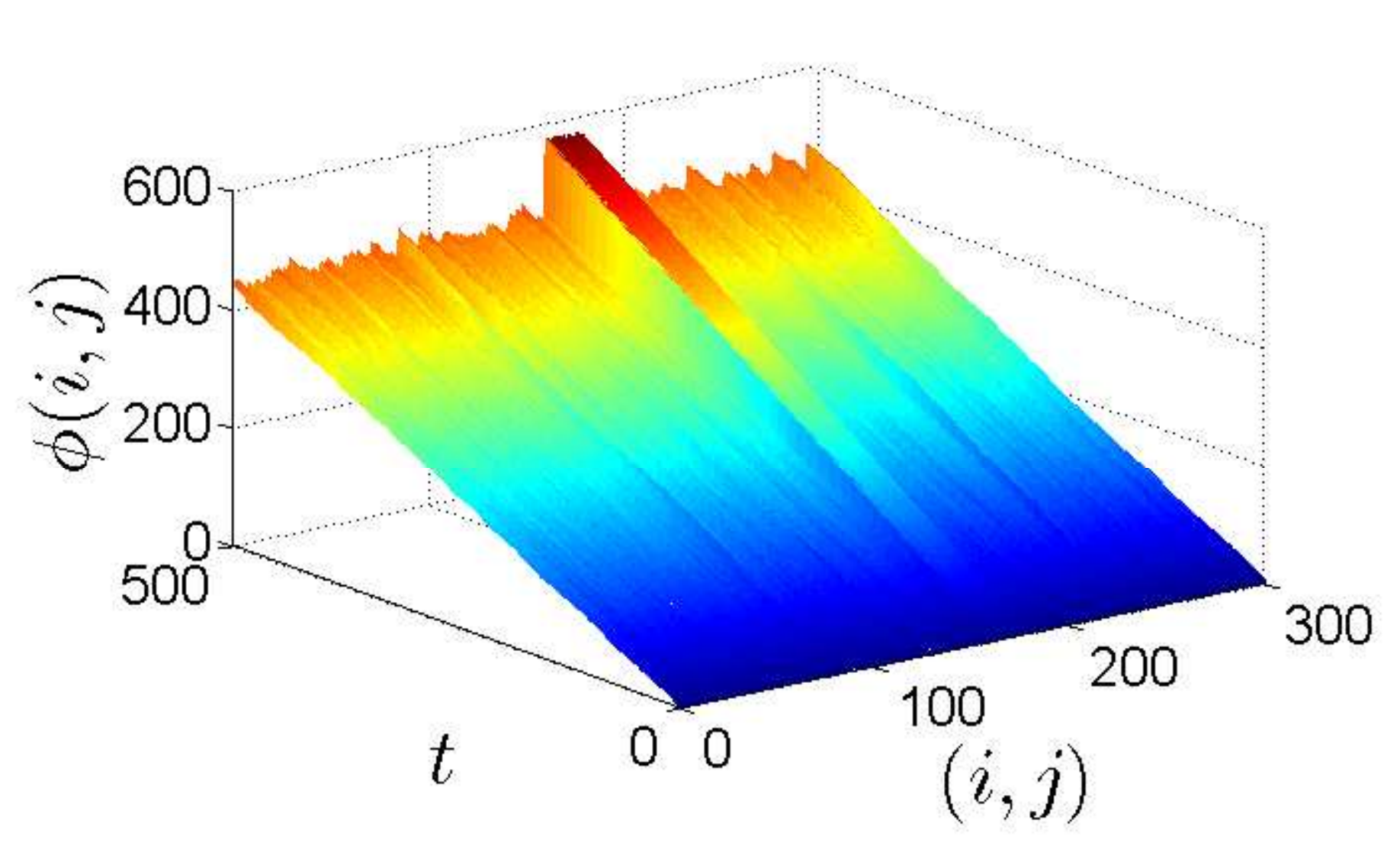}\label{fig:Artif02sphase}} \\
\subfloat[]
{\includegraphics[width = 8.0cm]{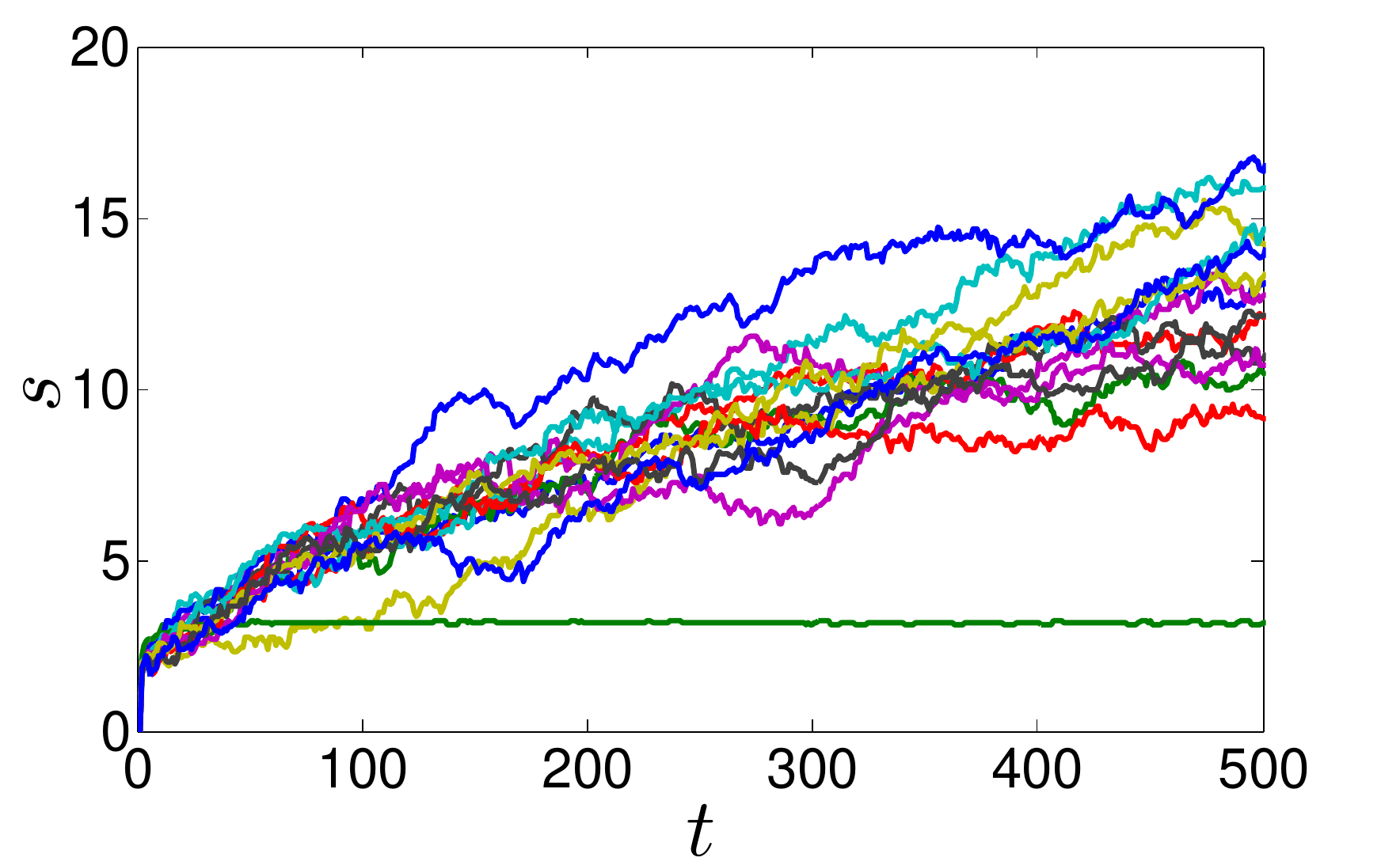}\label{fig:Artif01stdphase}}
\caption{Results of artificial image with medium contrast. (a) Oscillators behavior, $\sigma=0.40$. (b) Phase growth, $\sigma=0.40$. (c) Oscillators behavior, $\sigma=0.25$. (d) Phase growth, $\sigma=0.25$. (e) Time series of phase standard deviation of each object. $\sigma=0.25$.}
\end{figure}

The second experiment is carried out by using the artificial image shown by Fig. \ref{fig:Artif02src}. Like in Fig. \ref{fig:Artif01src} there are $15$ objects, and again the \textit{yellow} one (second line, forth column) is the salient. However, in this case there is less contrast between the salient object and the other objects. The free parameters were set as follows: $\sigma = 0.4$ and $\Delta\omega = 0.2$, the same values that were used in the previous experiment, so that we can observe what happens when there is less contrast in the scene. Figure \ref{fig:Artif02x} shows the behavior of $20$ randomly chosen oscillators (pixels) from each object, where each line corresponds to an oscillator. From lines $161$ to $180$, we can see that the oscillators corresponding to the salient object again form pattern indicating that  these oscillators are phase synchronized, however, in this case, phases of oscillators corresponding to other objects are not completely uncorrelated. Actually, some of the oscillators even show phase synchronized behavior, which means they are not sufficiently inhibited. Figure \ref{fig:Artif02phase} shows the oscillators phase growth along time, we see that not only the oscillators corresponding to the salient object but also the oscillators corresponding to other non-salient objects form  synchronized groups. Notice that we can prevent this behavior by setting a higher $\sigma$ value, compensating to the less contrast in the scene. In order to confirm it, we run the same experiment again with the same parameters, except for $\sigma$, which has its value lowered from $0.4$ to $0.25$. The results are shown by Figs. \ref{fig:Artif02sx} and \ref{fig:Artif02sphase}, and now we can observe that the salient object is able to completely inhibit other objects to be phase synchronized, as it is expected.

\begin{figure}
\centering
\subfloat[]
{\includegraphics[width = 8.0cm]{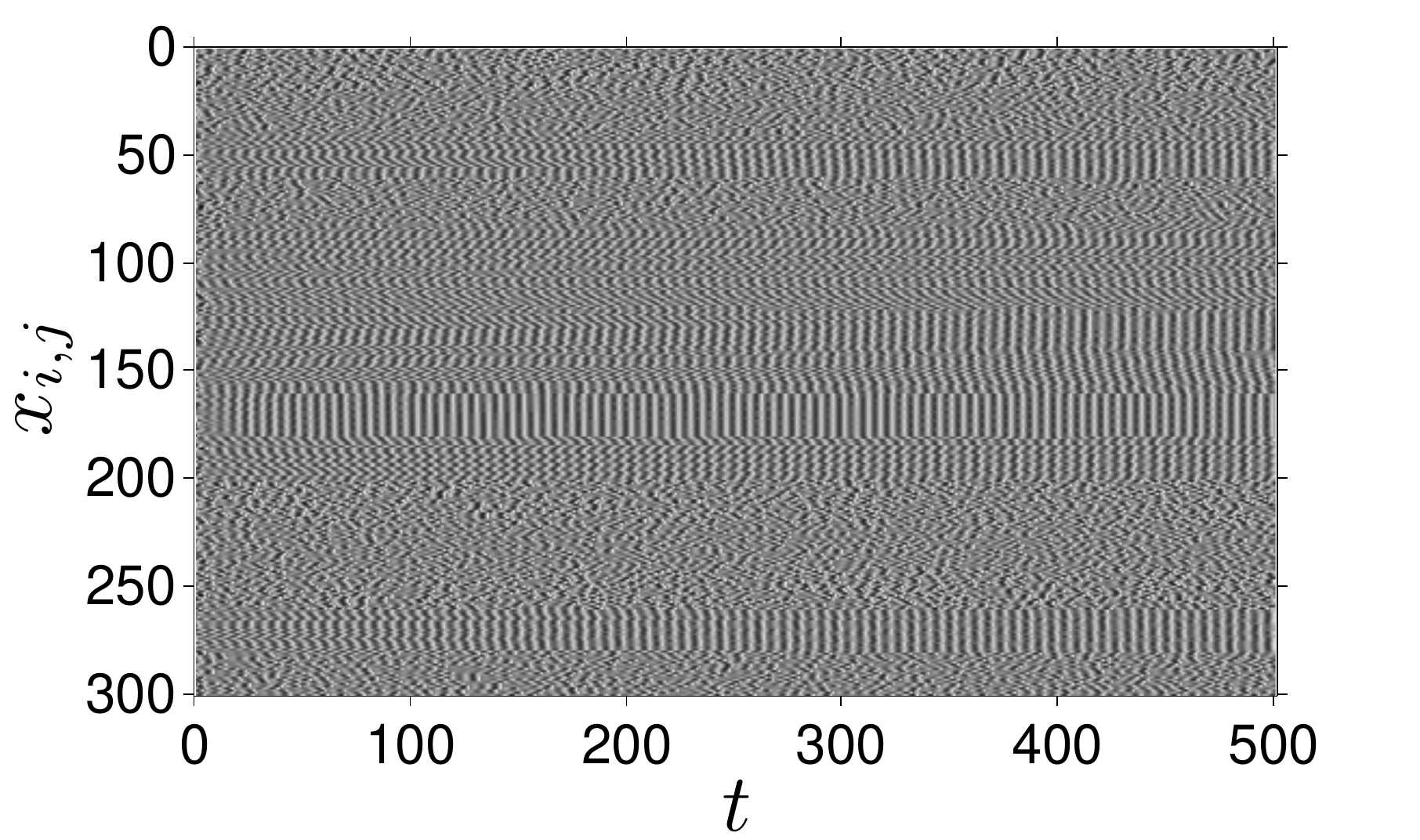}\label{fig:Artif03x}}
\subfloat[]
{\includegraphics[width = 8.0cm]{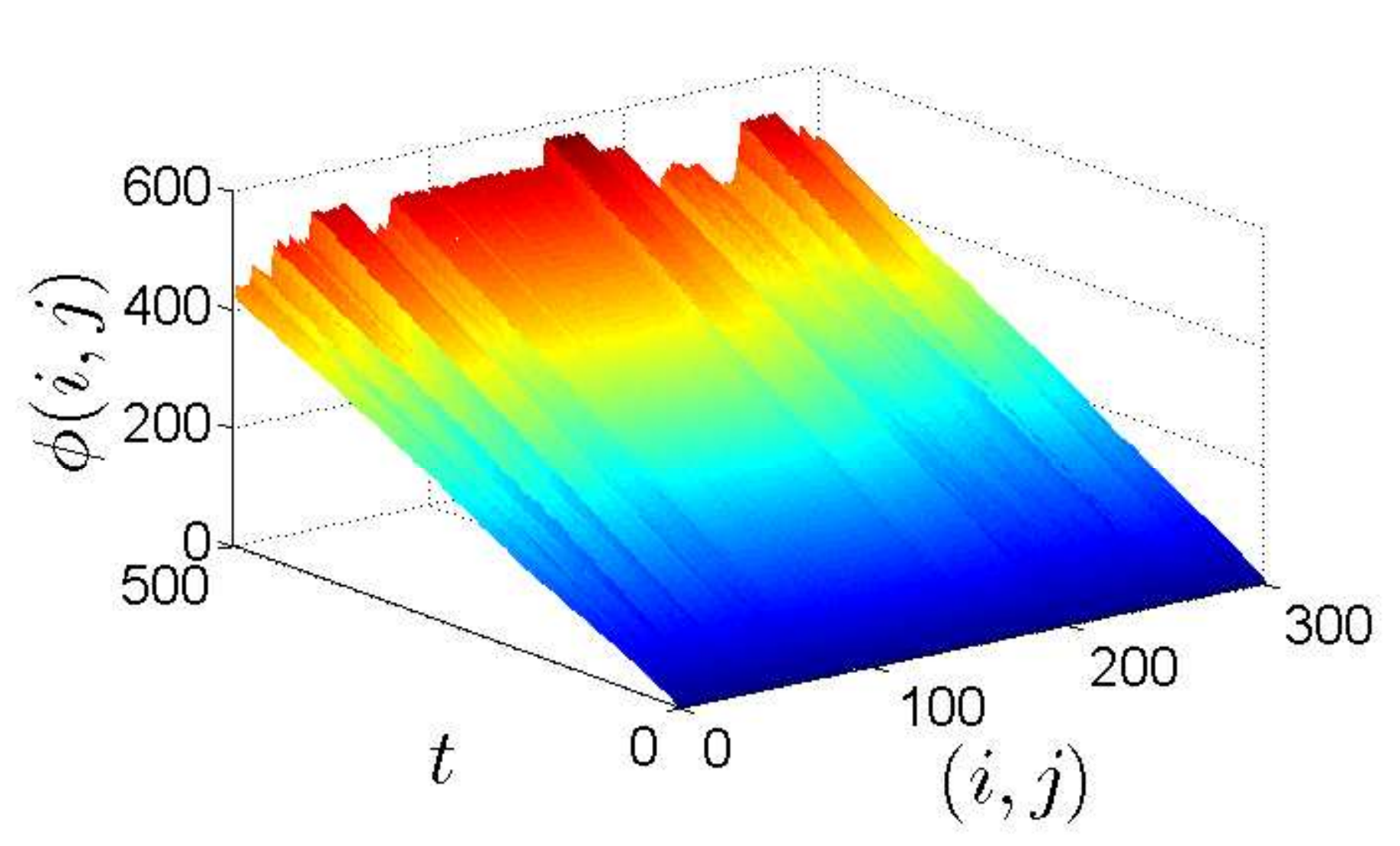}\label{fig:Artif03phase}} \\
\subfloat[]
{\includegraphics[width = 8.0cm]{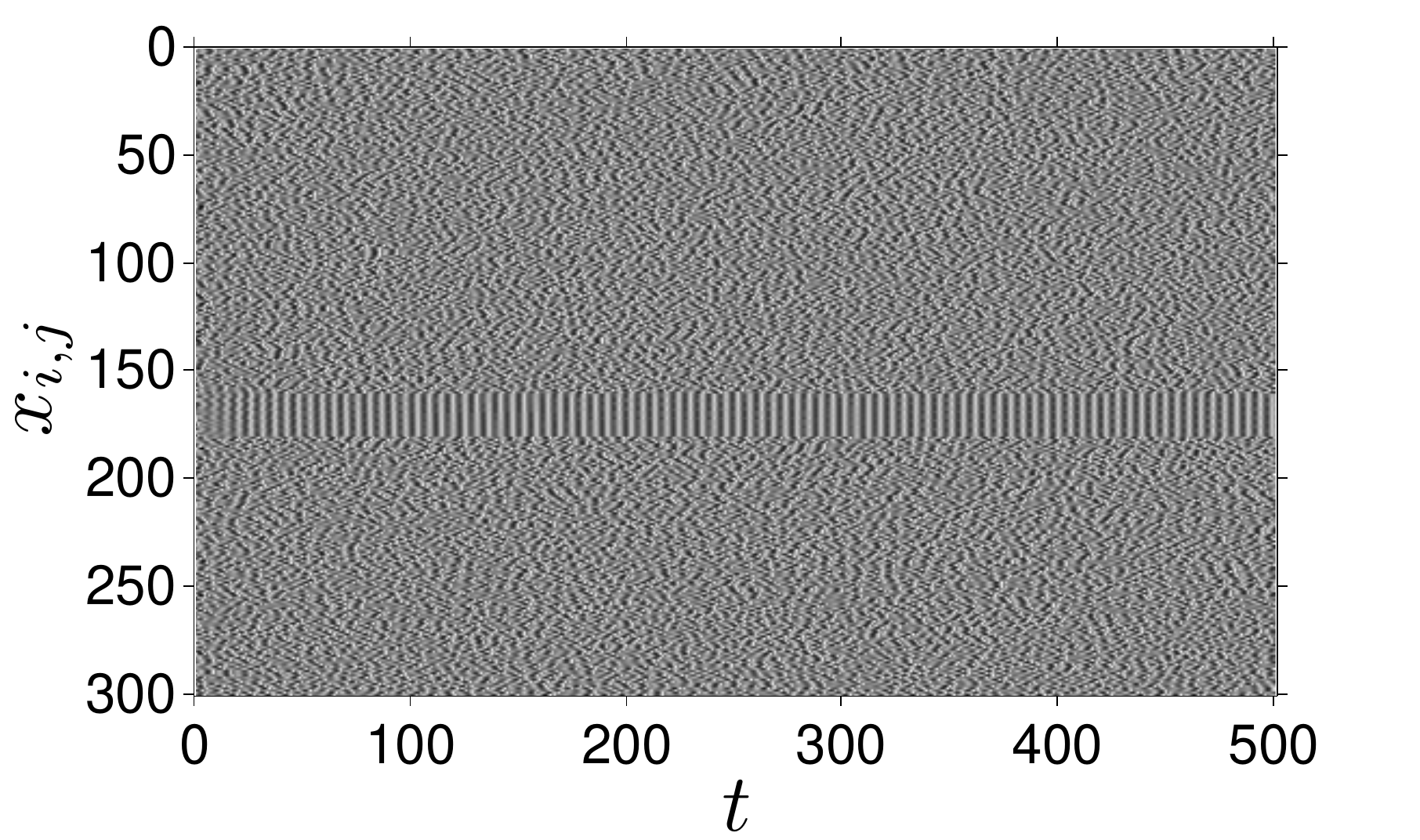}\label{fig:Artif03sx}}
\subfloat[]
{\includegraphics[width = 8.0cm]{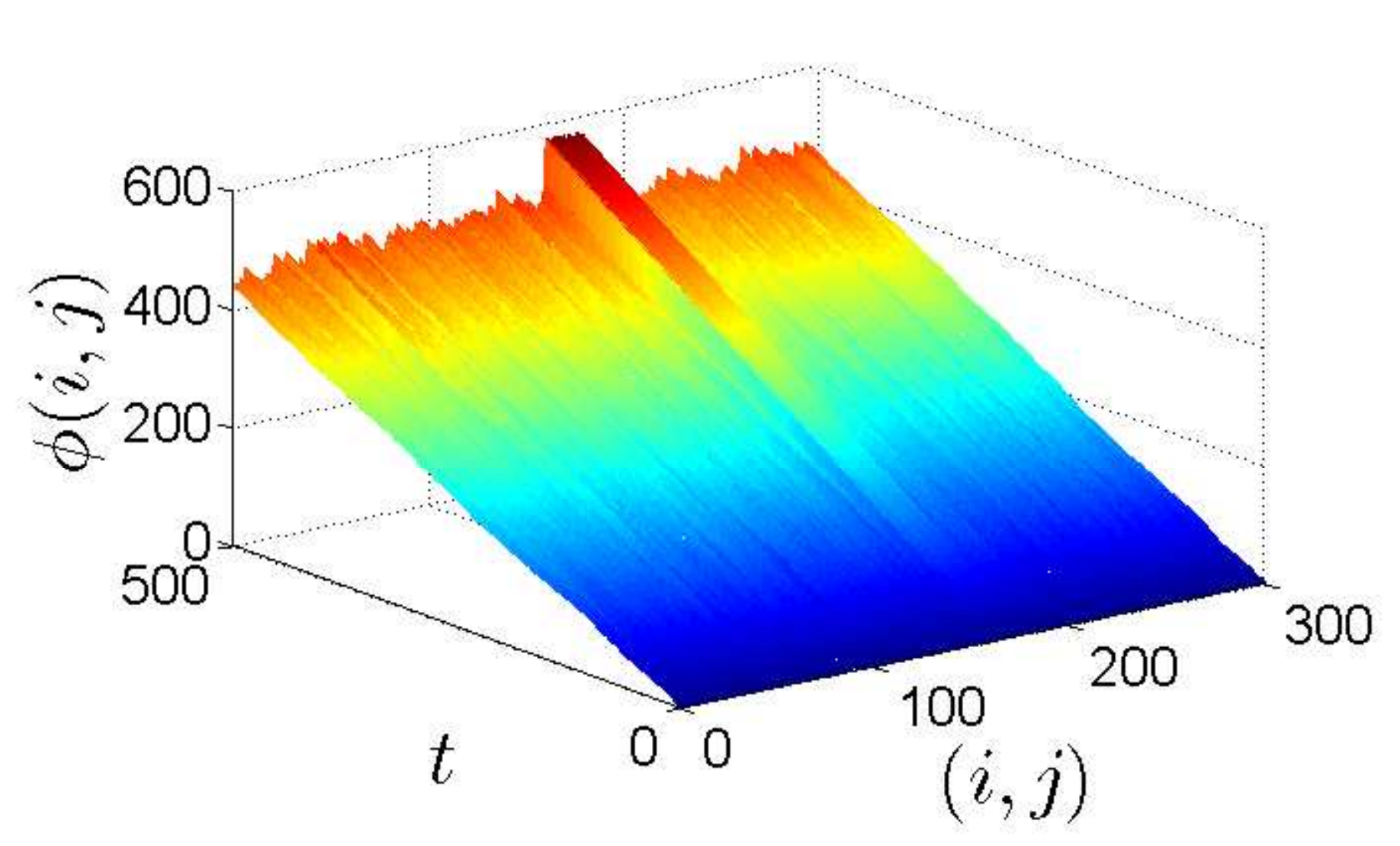}\label{fig:Artif03sphase}} \\
\subfloat[]
{\includegraphics[width = 8.0cm]{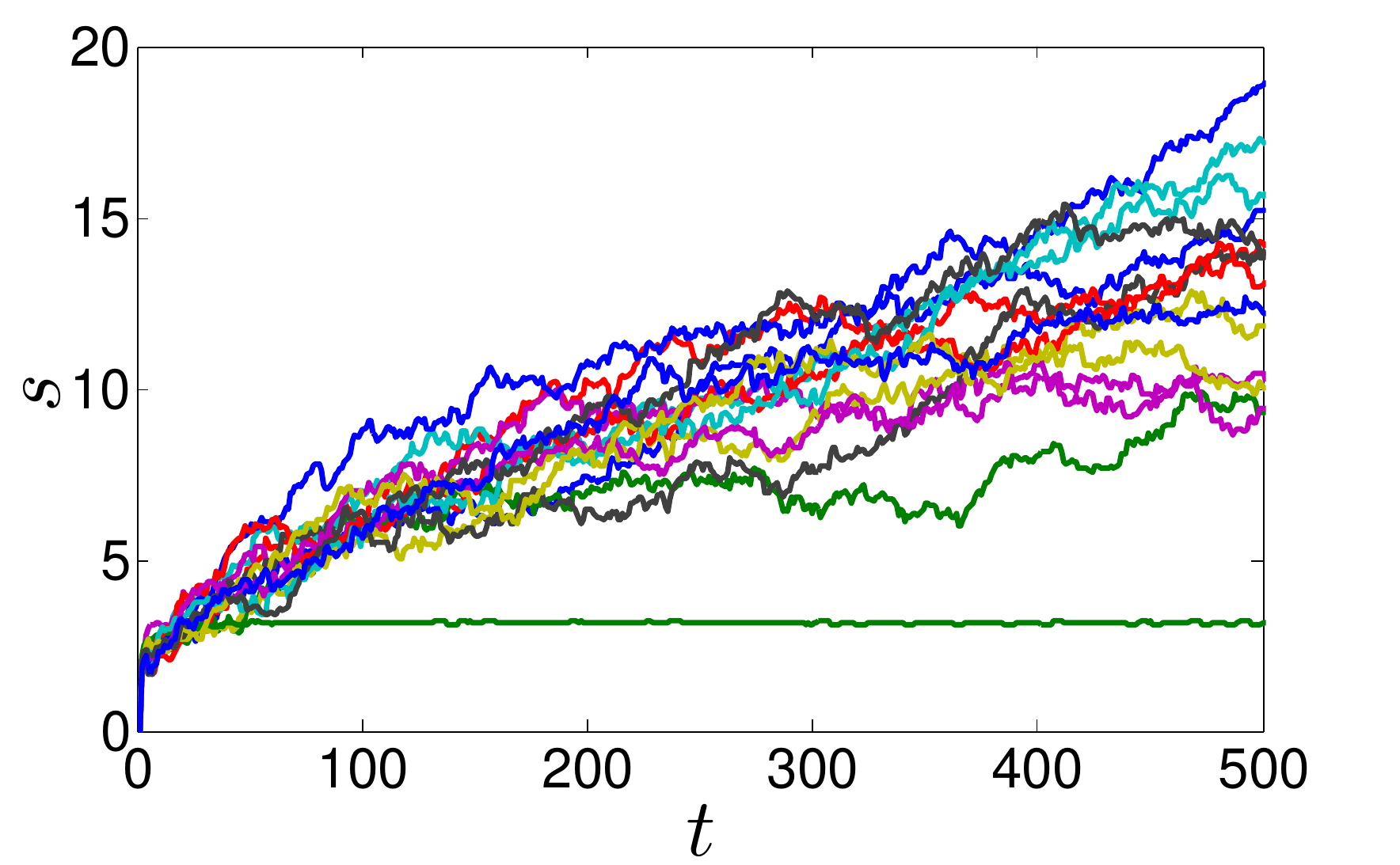}\label{fig:Artif01stdphase}}
\caption{Results of artificial image with high contrast. (a) Oscillators behavior, $\sigma=0.25$. (b) Phase growth, $\sigma=0.25$. (c) Oscillators behavior, $\sigma=0.10$. (d) Phase growth, $\sigma=0.10$. (e) Time series of standard deviation of phases for each object. $\sigma=0.10$.}
\end{figure}

Following to the third experiment, in Fig. \ref{fig:Artif03src} we have an artificial image with even less contrast, the \textit{yellow} object (second line, forth column) is again the salient object, but now there are bright surrounding objects of other colors. First, we perform simulation by using the parameters of the last experiment: $\Delta\omega = 0.2$ and $\sigma = 0.25$ and the results are shown by Figs. \ref{fig:Artif03x} and \ref{fig:Artif03phase}, where we can observe that $\sigma = 0.25$ is not sufficient to inhibit all the surrounding objects. Therefore, we lower $\sigma$ to $0.1$, and the new results are shown by Figs. \ref{fig:Artif03sx} and \ref{fig:Artif03sphase}, where we see that now the salient object is able to completely inhibit other ones.

\begin{figure}
\centering
\subfloat[]
{\includegraphics[width = 7.5cm]{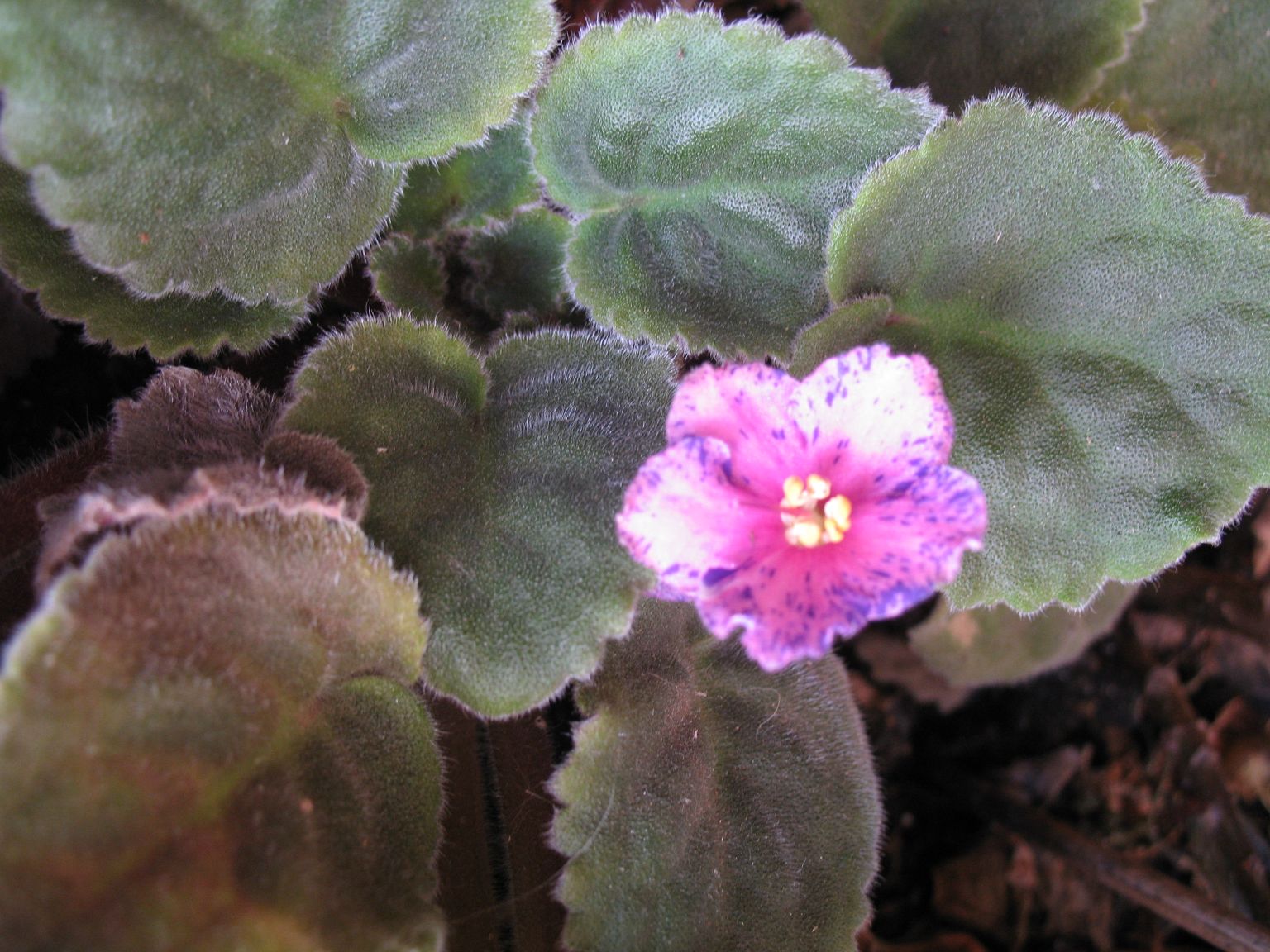} \label{fig:FlowerSrc}}
\subfloat[]
{\includegraphics[width = 8.0cm]{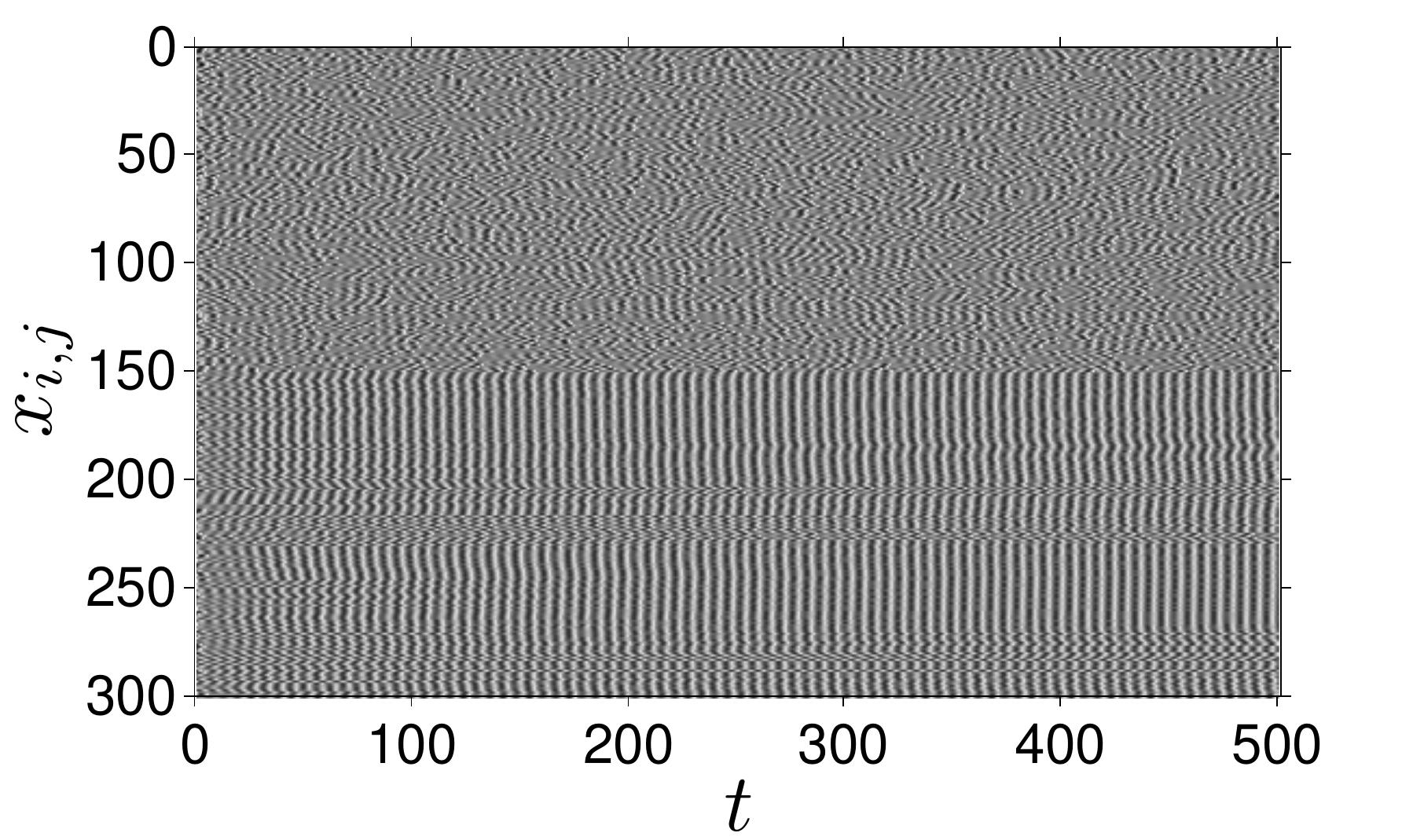}\label{fig:FlowerX}} \\
\subfloat[]
{\includegraphics[width = 8.0cm]{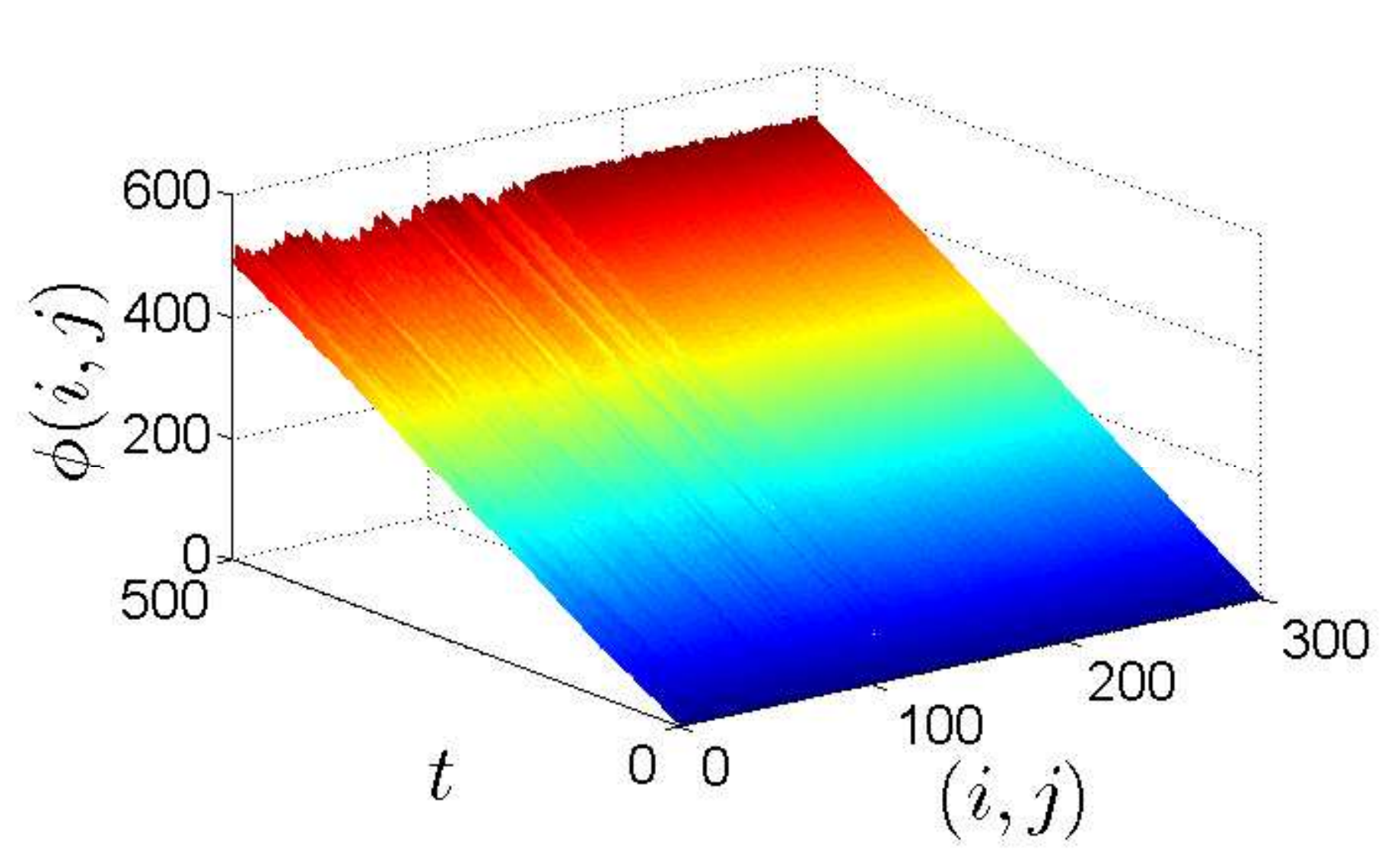}\label{fig:FlowerPhase}}
\subfloat[]
{\includegraphics[width = 8.0cm]{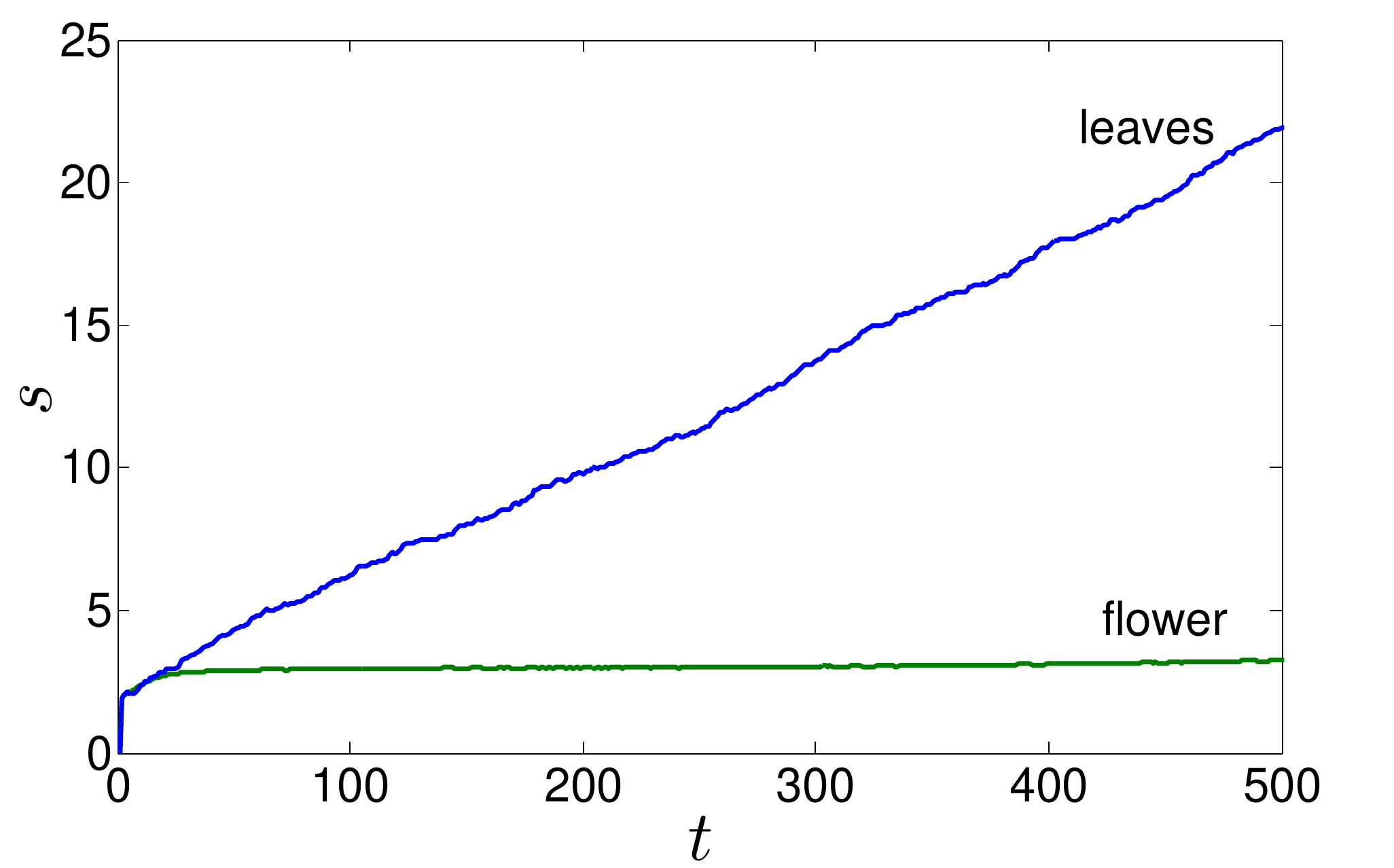}\label{fig:FlowerStdPhase}} \\
\subfloat[]
{\includegraphics[width = 8.0cm]{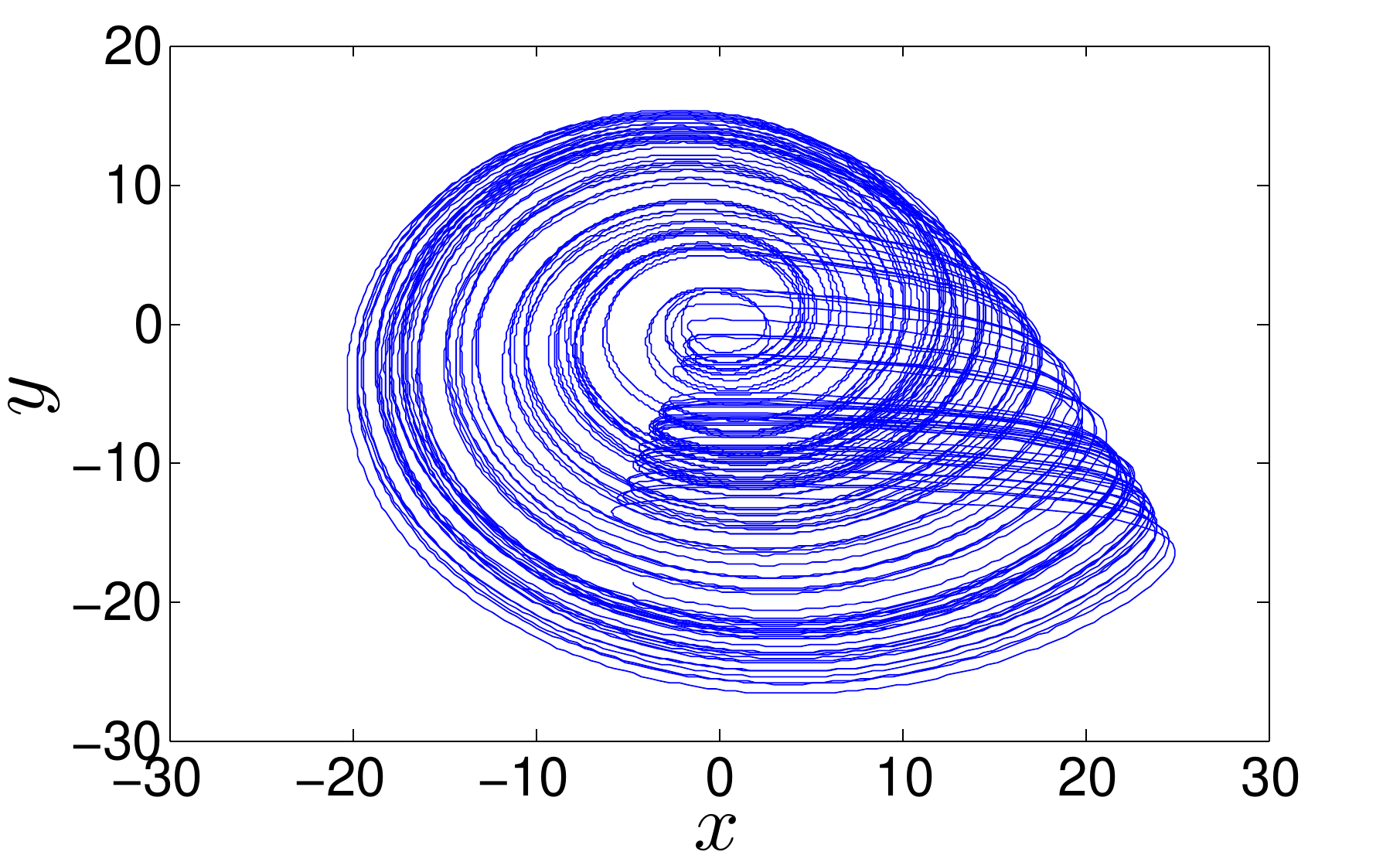}\label{fig:FlowerXYNonSync}}
\subfloat[]
{\includegraphics[width = 8.0cm]{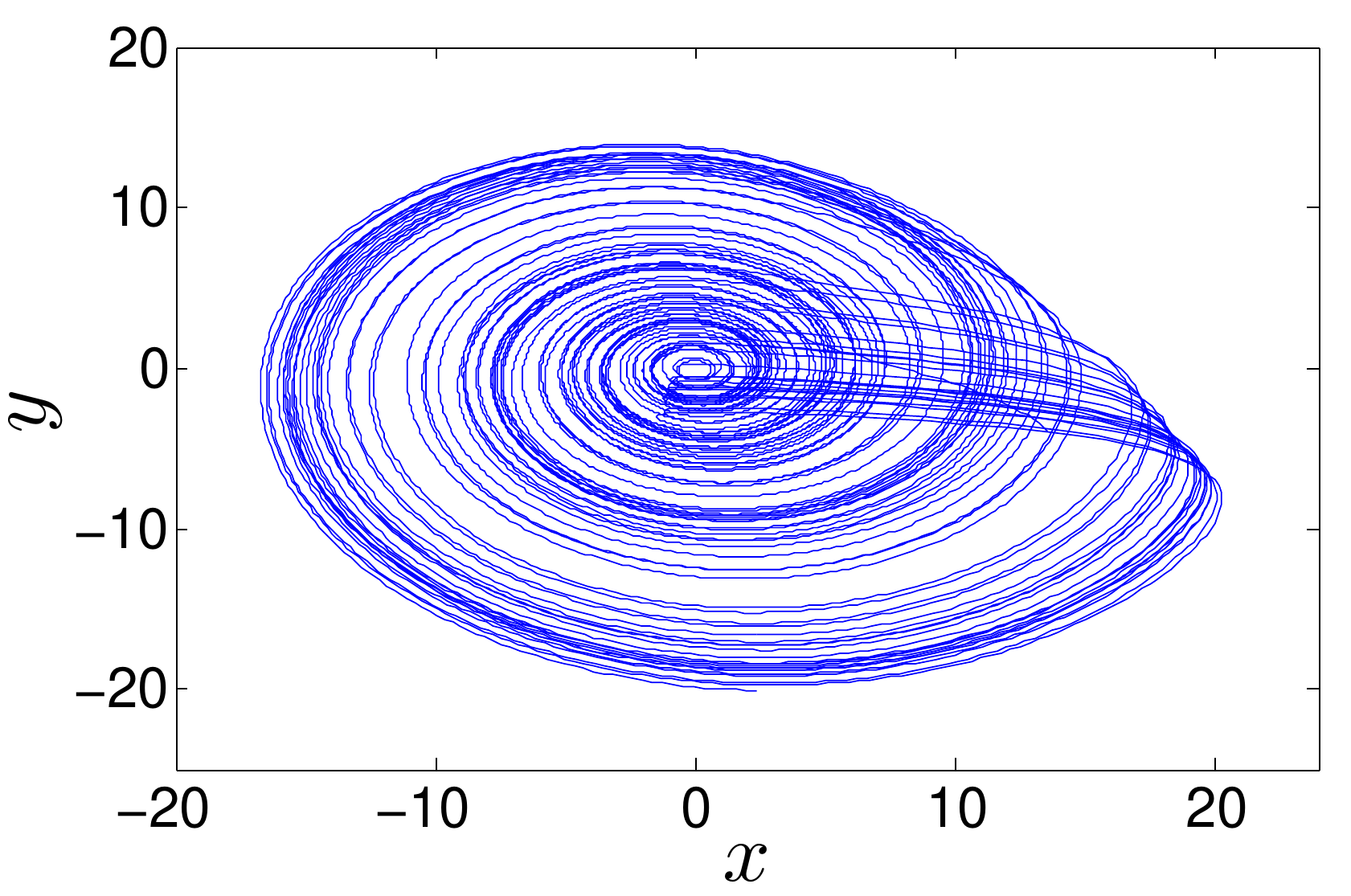}\label{fig:FlowerXYSync}}
\caption{Real Image - ``Flower'' (a) Source image. (b) Oscillators behavior. (c) Phase growth. (d) Time series of phase standard deviation of each object.
(e) Phase portrait of a randomly selected oscillator in background objects. (f) Phase portrait of a randomly selected oscillator in salient object.}
\end{figure}

The next experiment is performed by using a real-world image shown by Fig. \ref{fig:FlowerSrc}. The free parameters are set as follows: $\sigma = 0.5$ and $\Delta\omega = 0.02$. Figure \ref{fig:FlowerX} shows the behavior of $300$ randomly chosen oscillators (pixels) from the image, so the first $150$ lines correspond to the ``leaves'' and the other $150$ lines correspond to the ``flower''. Figures \ref{fig:FlowerPhase} and  \ref{fig:FlowerStdPhase} show that phase synchronization occurs among oscillators representing the object ``flower'', while no phase synchronization is observed among other oscillators. Thus, the object ``flower'' is selected, which agrees to our visual inspection. Again, Figs. \ref{fig:FlowerXYNonSync} and \ref{fig:FlowerXYSync} show that oscillators are still chaotic after the synchronization process settles down.

\begin{figure}
\centering
\subfloat[]
{\includegraphics[width = 6cm]{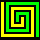} \label{fig:SpiralSrc}}  \\
\subfloat[]
{\includegraphics[width = 8.5cm]{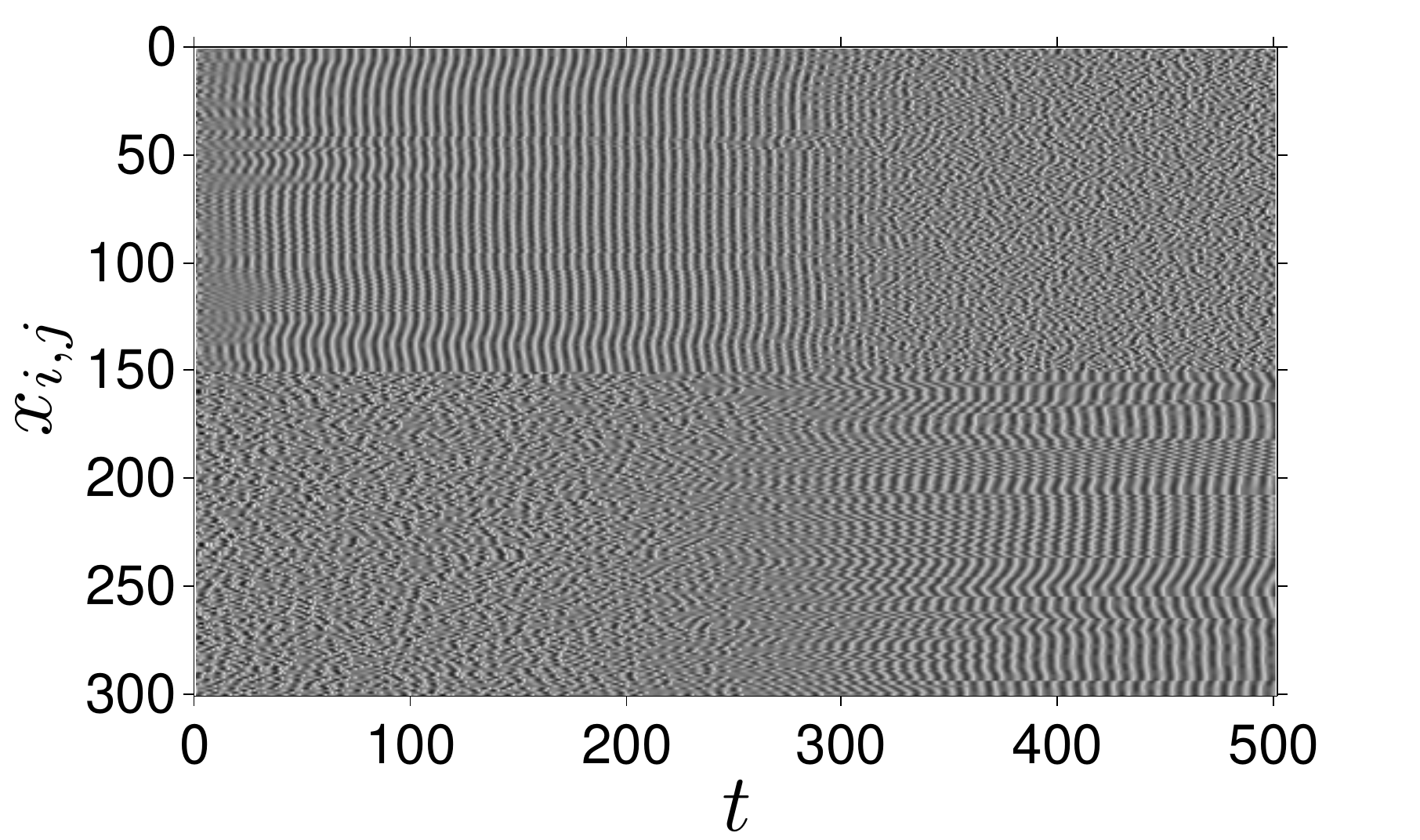}\label{fig:SpiralX}} \\
\subfloat[]
{\includegraphics[width = 8.0cm]{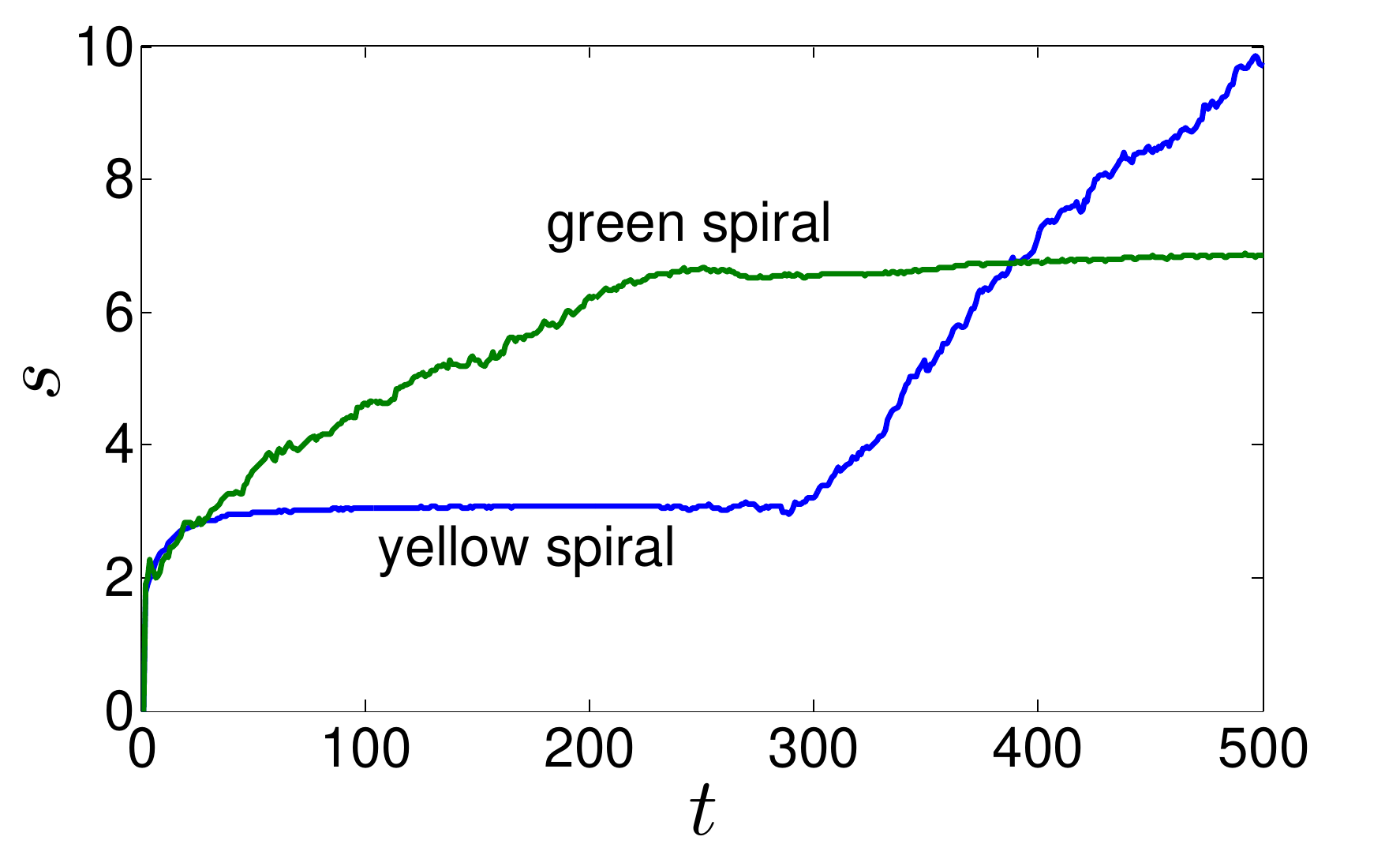}\label{fig:SpiralPhase}}
\caption{Artificial image - ``Spirals'' (a) Source image. (b) Oscillators behavior. (c) Time series of phase standard deviation of each object. }
\end{figure}

Now we present some experiments by using the shift mechanism to change the focus of attention from one object to another. In Fig. \ref{fig:SpiralSrc}, we show an artificial image with two spirals. The free parameters are set as follows: $\sigma = 0.3$ and $\Delta\omega = 0.2$. Figure \ref{fig:SpiralX} shows the behavior of $150$ randomly chosen oscillators (pixels) from each object, where each line corresponds to an oscillator. From lines $1$ to $150$ we can see that the oscillators corresponding to the \textit{yellow} object are the first group to be phase synchronized. After some time, they lose their synchronization and the phase synchronization of the second group (lines $151$ to $300$) emerges. Figure \ref{fig:SpiralPhase} shows the standard deviations of phase growth of the two groups of oscillators, respectively. We see that when an object is salient, the phase standard deviation of its corresponding oscillators forms a plateau indicating that they are phase synchronized. Meanwhile, when an object is not salient, the phase standard deviation of its corresponding oscillators grows continuously.

\begin{figure}
\centering
\subfloat[]
{\includegraphics[width = 7.5cm]{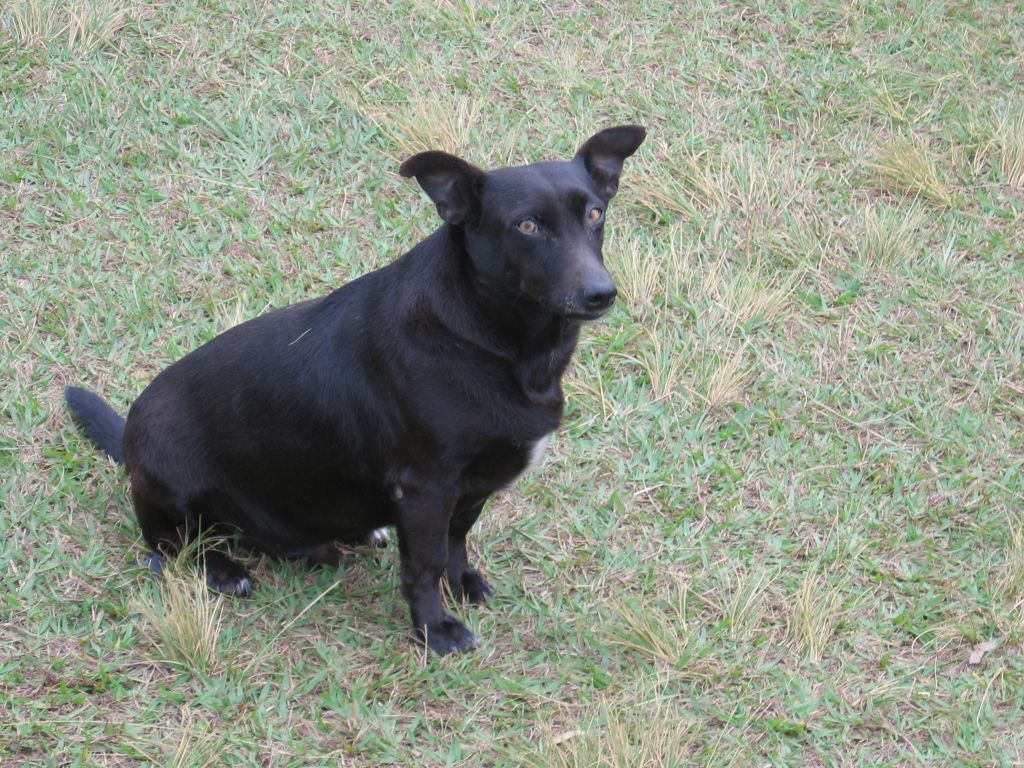} \label{fig:DogSrc}}  \\
\subfloat[]
{\includegraphics[width = 8.5cm]{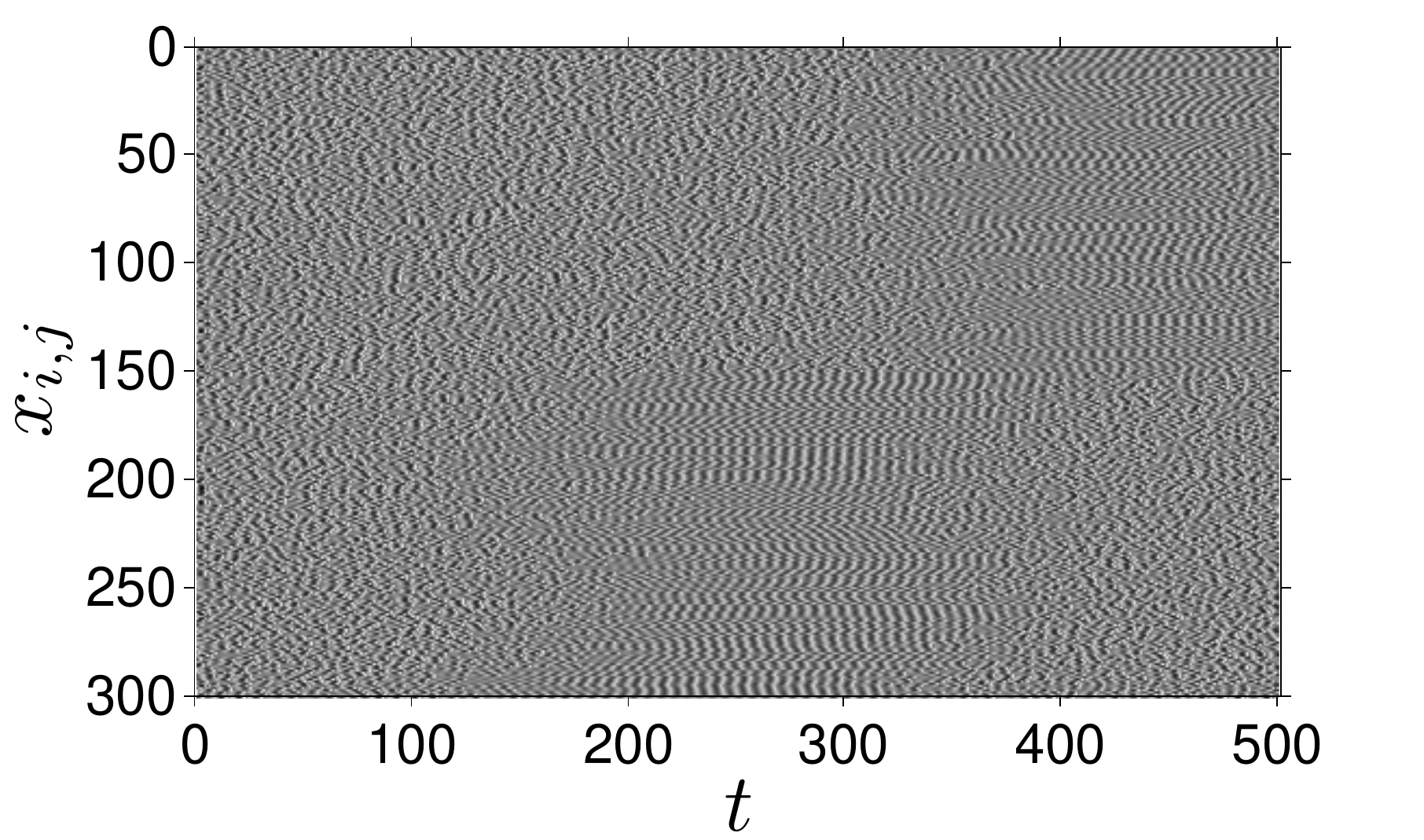}\label{fig:DogX}} \\
\subfloat[]
{\includegraphics[width = 8.0cm]{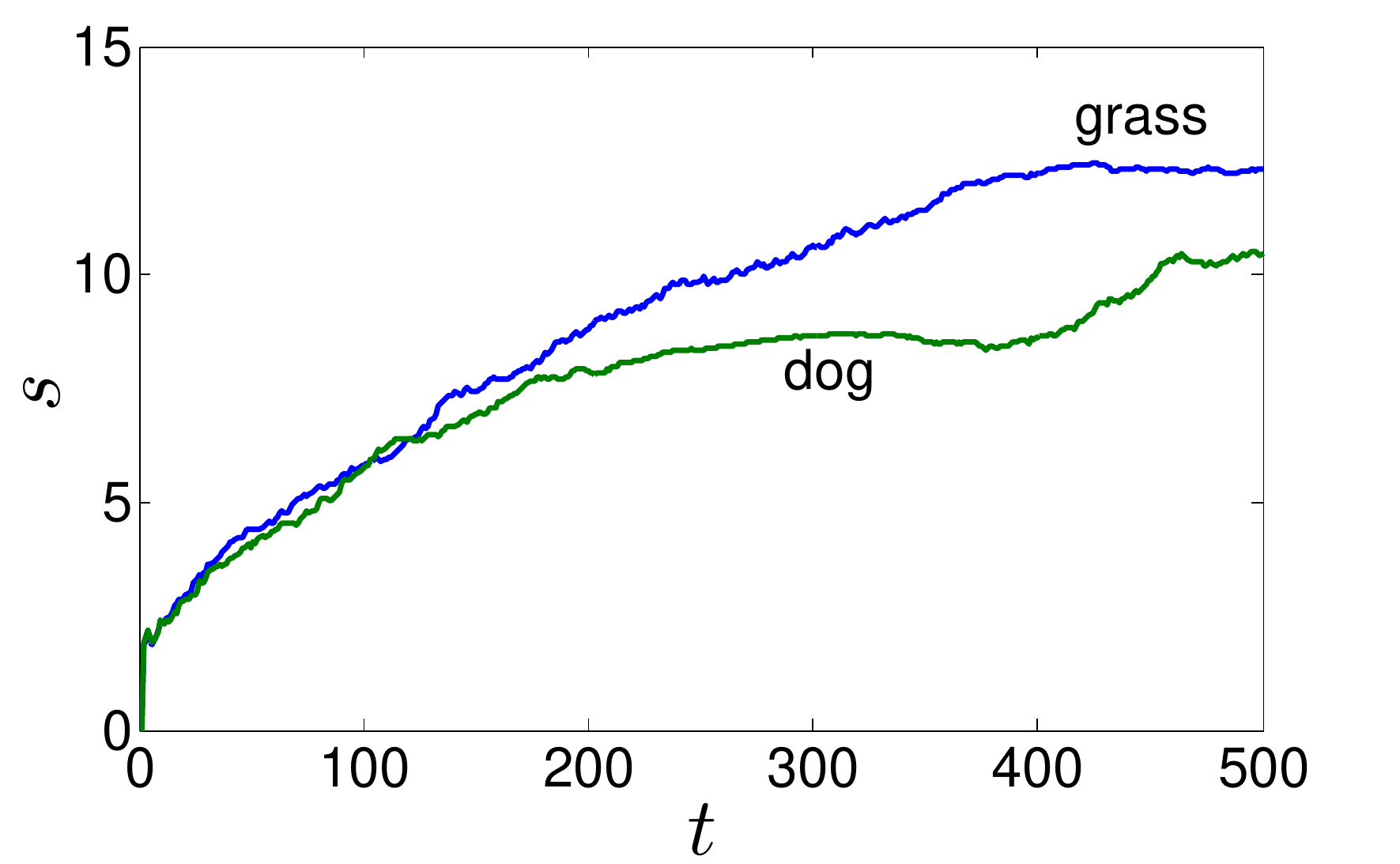}\label{fig:DogPhase}}
\caption{Real image - ``Dog'' (a) Source image. (b) Oscillators behavior. (c) Time series of phase standard deviation of each object. }
\end{figure}

Now we apply the shift mechanism to the real-world image shown by Fig. \ref{fig:DogSrc}. In this figure, the salient object is the ``dog'', which has an irregular form and is surrounded by the irregular background ``grass''. The free parameters were set to: $\sigma = 0.1$ and $\Delta\omega = 0.02$. Figure \ref{fig:DogX} shows the behavior of $300$ randomly chosen oscillators (pixels) from the image. The first $150$ lines correspond to the ``grass'' and the other $150$ lines correspond to the ``dog''. We see that Fig. \ref{fig:DogX} can be divided into three stages: the first stage is from time zero to about $200$, no object is salient (no phase synchronization is observed); the second stage is from about time $200$ to about time $350$, oscillators corresponding to object ``dog'' are phase synchronized and thus the object ``dog'' is salient at this stage; the third stage is from about time $350$ to time $500$, oscillators corresponding to object ``dog'' lose synchronization and, at the same time, oscillators corresponding to object ``grass'' become synchronized. Figure \ref{fig:DogPhase} shows the standard deviation of oscillators phase growth through time. Again, we see that the system delivers the focus of attention firstly to the ``dog'', after some time, shifting to the object ``grass''.

In this work the free parameters were set empirically according to the input image. The $\sigma$ parameter defines how much the salient object will inhibit the other objects in the scene, as demonstrated in the second and third experiments. When the salient object is homogenous there is no harm in setting a lower value for $\sigma$. However, when the scene presents a larger and more heterogenous salient object, the $\sigma$ parameter has to be chosen carefully in order to prevent the salient object from inhibiting other parts of itself if this is not a desirable behavior. The other parameters are also sensitive, and the positive coupling strength $k^{+}_{max}$ cannot be set too low because it would not be able to keep the salient object synchronized, and neither it can be set too high because it would delay or even prevent the inhibition of non-salient objects. The $\Delta\omega$ parameter defines how much the oscillators will differ from each other based on their respective absolute contrast, and consequently it will also define their different oscillation speeds and how much their phase growth tends to be different when they are not coupled. Therefore, $\Delta\omega$ has to be carefully chosen, as it directly affects the synchronization of the salient object and the inhibition of the non-salient ones. As future work, we intend to develop some mechanism to optimize these parameters automatically.

\section{Conclusions}

This paper presented a visual selection mechanism based on a network composed of chaotic R\"ossler oscillators, taking advantage of its phase synchronization behavior. This mechanism can be seen as part of a visual attention system, which is responsible for selecting one object of interest  from an input image at each time, while keeping the non-salient, or less salient, objects unsynchronized.

Oscillatory networks have been applied to solve problems of image segmentation, auditory signal segregation, feature binding, and object selection. This kind of models requires two mechanisms: synchronization to group each object and desynchronization to distinguish one object form another. Network of coupled periodical oscillators has advantage to easily synchronize each group of oscillators. But it has difficulty to discriminate different objects due to accident coincidence of various synchronized trajectories. Chaos and chaotic synchronization is a suitable solution to this problem. This is because a group of chaotic oscillators can be synchronized and at the same time all groups of oscillators can be easily distinguished following the high sensitivity to initial conditions and the dense properties of chaos \citep{Hansel1992,Zhao2000,Zhao2001}. However, complete chaotic synchronization of a large amount of locally coupled oscillators (like in image processing applications) requires a strong coupling strength, which may leads the system to diverge to infinity. On the other hand, the coupling strength required for chaotic phase synchronization is much smaller, thus the system is safe. Moreover, chaotic phase synchronization is more robust than chaotic complete synchronization or periodic synchronization. This feature can be seen from the simulation results performed on real-world images shown by Figs. \ref{fig:FlowerSrc} and \ref{fig:DogSrc}, where the pixels within the ``flower'' object or within the ``grass'' object vary in a certain range. In these cases, the oscillators representing the ``flower'' object or the ``grass'' object may not be completely synchronized but can be phase synchronized. This is a desirable feature in image segmentation and object selection. Moreover, chaotic phase synchronization is usually observed in nonidentical systems. This is also desirable in engineering applications, since real systems are rarely identical.

\section{Acknowledgements}

This work was supported by the State of S\~ao Paulo Research Foundation (FAPESP) and the Brazilian National Council of Technological and Scientific Development (CNPq)

%
%
%

\bibliographystyle{apalike}
\bibliography{ijcnn2009extended}

\end{document}